\documentclass[11pt]{article}

\usepackage[final]{acl}

\usepackage{times}
\usepackage{latexsym}
\usepackage{booktabs}
\usepackage{amsmath}
\usepackage{amsfonts}

\usepackage[T1]{fontenc}

\usepackage[utf8]{inputenc}

\usepackage{microtype}

\usepackage{inconsolata}

\usepackage{graphicx}
\usepackage{titletoc}
\usepackage{colortbl}
\usepackage{ulem}
\usepackage{multirow}

\newcommand\DoToC{%
  \startcontents
  \printcontents{}{1}{\textbf{\large Contents of Appendix}\vskip3pt\hrule\vskip5pt}
  \vskip3pt\hrule\vskip5pt
}

\title{FlowMind: Execute-Summarize for Structured Workflow Generation from LLM Reasoning}

\author{
  \textbf{Yihao Liu}\textsuperscript{1,3} \quad \textbf{Ziyun Zhang}\textsuperscript{1,3} \quad \textbf{Zile He}\textsuperscript{2} \quad \textbf{Huaqian Cai}\textsuperscript{1,3}\thanks{Corresponding author: \texttt{caihq@pku.edu.cn}}
  \\
  \textsuperscript{1}Peking University
  \\
  \textsuperscript{2}East China University of Technology
  \\
  \textsuperscript{3}National Key Laboratory of Data Space Technology and System
}

\begin{document}
\maketitle
\begin{abstract}
LLMs can solve complex tasks through reasoning and tool use, but accurately translating these solutions into structured workflows remains challenging. We model workflows as sequences of tool use and reformulate the problem as designing a mechanism that can both solve tasks and reliably construct workflows. Prior approaches that build workflows during execution often suffer from inaccuracies due to interference between the two processes. We propose an execute–summarize framework that decouples task execution from workflow construction: the model first completes the task using available tools, then independently reconstructs a structured workflow from execution traces. This separation improves workflow accuracy and robustness. We introduce FlowBench and show through extensive experiments that our approach outperforms existing methods, providing a reliable paradigm for grounding free-form LLM reasoning into structured workflows.
\end{abstract}

\section{Introduction}
\begin{figure}[ht!]
    \centering
    \includegraphics[width=\linewidth]{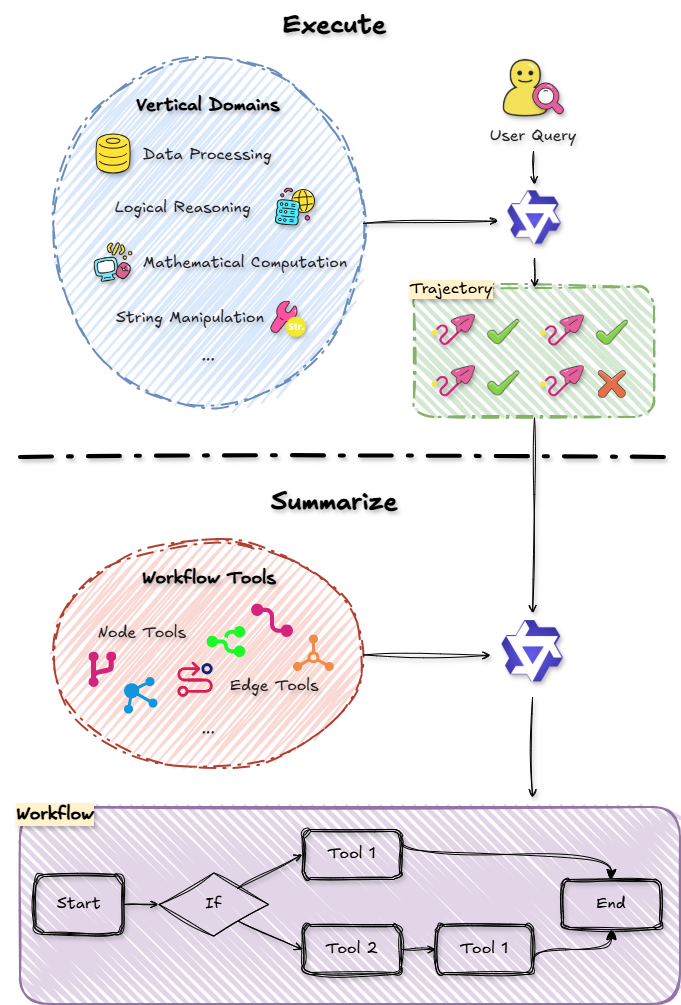}
    \caption{Overview of the \textbf{Execute-Summarize} design in FlowMind. The Execute phase uses domain tools to generate task trajectories from user queries, while the Summarize phase converts these trajectories into a structured workflow via workflow tools.}
    \label{fig:workflow_summary_pipeline}
\end{figure}

Large Language Models (LLMs) have demonstrated strong capabilities in solving complex tasks through multi-step reasoning and interaction with external tools, as exemplified by Toolformer\cite{schick2023toolformer} and ReAct\cite{yao2022react}. By extending beyond pure text generation, tool-augmented LLMs support a wide range of applications, including automated task execution, intelligent agents, and decision support systems, such as HuggingGPT \cite{shen2023hugginggpt}, SciAgent \cite{ma2024sciagent}, and UI-TARS \cite{qin2025ui}. However, in many real-world settings, successfully completing a task is not sufficient on its own. Tool-augmented LLMs often face challenges arising not only from under-specified user queries or unavailable tools \cite{trevino2025benchmarking}, but also from limited transparency in their reasoning process, difficulties in reproducing results, uncertainties in dynamic environments.\cite{trevino2025benchmarking, liu2024uncertainty, hochlehnert2025sober}

In practical settings, simply completing a task is often insufficient. Ensuring correctness, reliability, and reproducibility demands well-defined, deterministic processes rather than relying on agents with inherently uncertain behavior \cite{qiu2025blueprintfirstmodelsecond}. Consequently, tool-augmented LLM outputs must be translated into structured workflows—explicit sequences of tool invocations with clearly defined dependencies and execution order \cite{liang2025dataflowllmdrivenframeworkunified}. Structuring LLM reasoning in this way enhances interpretability, facilitates debugging and auditing, and enables reproducibility, reuse, and downstream automation \cite{zeng2025airepranalystinspectorframeworkevaluating}. As a result, reliably converting free-form LLM reasoning and tool use into structured workflows has become a critical challenge for real-world applications.

A common approach is to construct workflows during task execution, requiring the model to reason about the task while simultaneously producing a structured representation of its actions. However, this coupling introduces significant challenges. On the one hand, the cognitive burden of maintaining workflow structure can interfere with task reasoning, leading to suboptimal decisions or incorrect tool usage.\cite{adapala2025cognitive, kaiser2025cogniload, shang2025united} On the other hand, execution-time workflow construction is prone to errors such as missing steps, incorrect ordering, or inconsistent arguments, especially in long-horizon or tool-intensive tasks.\cite{wang2023stepsbenchmarkorderreasoning, He_2025} These issues make the resulting workflows unreliable, limiting their usefulness in real-world systems.

In this work, we model workflows as sequences of workflow tool invocations and reformulate the problem as jointly solving tasks and constructing accurate workflows. We identify a key limitation of prior approaches in their tight coupling of task execution and workflow generation, which often degrades both reasoning quality and workflow correctness. To address this limitation, we propose an \textbf{execute-summarize} framework (Figure~\ref{fig:workflow_summary_pipeline}), in which the model first focuses exclusively on completing the task using available tools without imposing structural constraints, and then independently reconstructs a structured workflow by summarizing the resulting execution traces. This explicit decoupling enables the model to fully leverage its reasoning capacity during execution, while producing more accurate, consistent, and reliable workflows in a dedicated summarization stage.

To systematically evaluate workflow construction, we introduce FlowBench, a benchmark designed to assess both task success and workflow accuracy across diverse tool-use scenarios. Extensive experiments show that the proposed execute–summarize framework consistently improves workflow correctness and maintains competitive task performance.

Overall, our contributions are threefold:
\begin{itemize}
  \item We identify the \textbf{cognitive burden} induced by jointly performing task execution and workflow construction as a major source of errors in existing approaches.
  \item We propose an \textbf{Execute-Summarize} framework that decouples execution from workflow generation, leading to more reliable and consistent workflows.
  \item We introduce \textbf{FlowBench}, a benchmark for systematic evaluation of workflow quality, and provide extensive empirical results demonstrating the effectiveness of our approach.
\end{itemize}

\section{Related Work}
\subsection{Workflow Representation}
Prior work adopts diverse representations for workflows in LLM-based systems. Some approaches model workflows as executable code or programs, offering strong expressiveness but requiring strict syntactic correctness.\cite{zhang2024aflow, hu2024automated, xu2024llm4workflow} Others rely on predefined templates or schemas, which improve validity but limit flexibility and generalization.\cite{qiu2025blueprintfirstmodelsecond, fagnoni2025opus, wu2024stateflow} Some approaches represent workflows using structured JSON specifications, prompting LLMs to generate schema-constrained JSON objects that explicitly encode task steps, dependencies, and control flow. This design improves robustness and machine interpretability while retaining more flexibility than rigid templates.\cite{kang2025scalingllmplanningnl2flow, li2024autoflowautomatedworkflowgeneration, wu2024stateflow, ayala2025finetuneslmpromptllm} In contrast, many modern agent systems implicitly represent workflows through tool abstractions, where each step corresponds to a tool invocation with structured inputs and outputs.\cite{yao2022react, wu2024autogen, hong2023metagpt}

In this work, we explicitly adopt a tool-based representation of workflows, modeling them as structured sequences of tool calls. This choice is motivated by the fact that contemporary large language models are natively trained for tool invocation and multi-turn interaction, making tool-centric workflows a natural and well-aligned abstraction.\cite{yang2025qwen3technicalreport, kimiteam2025kimik2openagentic, zeng2025glm, qin2025ui} By grounding workflows in the same interface used during task execution, this representation balances executability, flexibility, and compatibility with existing LLM agent capabilities.

\subsection{Workflow Generation}
Existing workflow generation methods for LLM-based agent systems mainly focus on integrating workflow construction with task execution or relying on structured supervision, which can be roughly divided into three paradigms: in-execution workflow construction, template/rule-based generation, and learning-based end-to-end generation . These approaches differ in technical routes but all face challenges in balancing task performance and workflow reliability .

In-execution workflow construction, the most common paradigm, interleaves task reasoning, tool invocation, and workflow structuring. Early works like ReAct\cite{yao2022react} and Toolformer\cite{schick2023toolformer} implicitly encode workflows through action sequences, while extended methods such as Plan-and-Execute add pre-planning stages, but both still couple execution with workflow construction.\cite{He_2025, erdogan2025plan} Template/rule-based methods leverage predefined schemas or rules to ensure workflow validity but suffer from poor generalization.\cite{qiu2025blueprintfirstmodelsecond, fagnoni2025opus, wu2024stateflow} Learning-based end-to-end methods fine-tune LLMs to generate structured workflows directly from queries, offering flexibility but requiring large annotated datasets and suffering from disconnection from actual execution dynamics.\cite{fan2024workflowllmenhancingworkfloworchestration, zhu2025adaptflowadaptiveworkflowoptimization, gschwind2025classifieraugmentedgenerationstructuredworkflow, lin2025typecompliantadaptationcascadesadapting, ayala2025finetuneslmpromptllm, fagnoni2025opus}

A common limitation across these paradigms is the lack of explicit decoupling between task execution and workflow construction, resulting in trade-offs between task performance and workflow accuracy. In contrast, our proposed execute–summarize framework addresses this by separating the two processes, leveraging execution traces to reconstruct workflows. This avoids the cognitive burden of in-execution structuring, inflexibility of templates, and execution disconnect of end-to-end methods, which is a direction rarely explored systematically in prior work.

\section{Methodology}
\begin{figure*}[t]
    \centering
    \includegraphics[width=\textwidth]{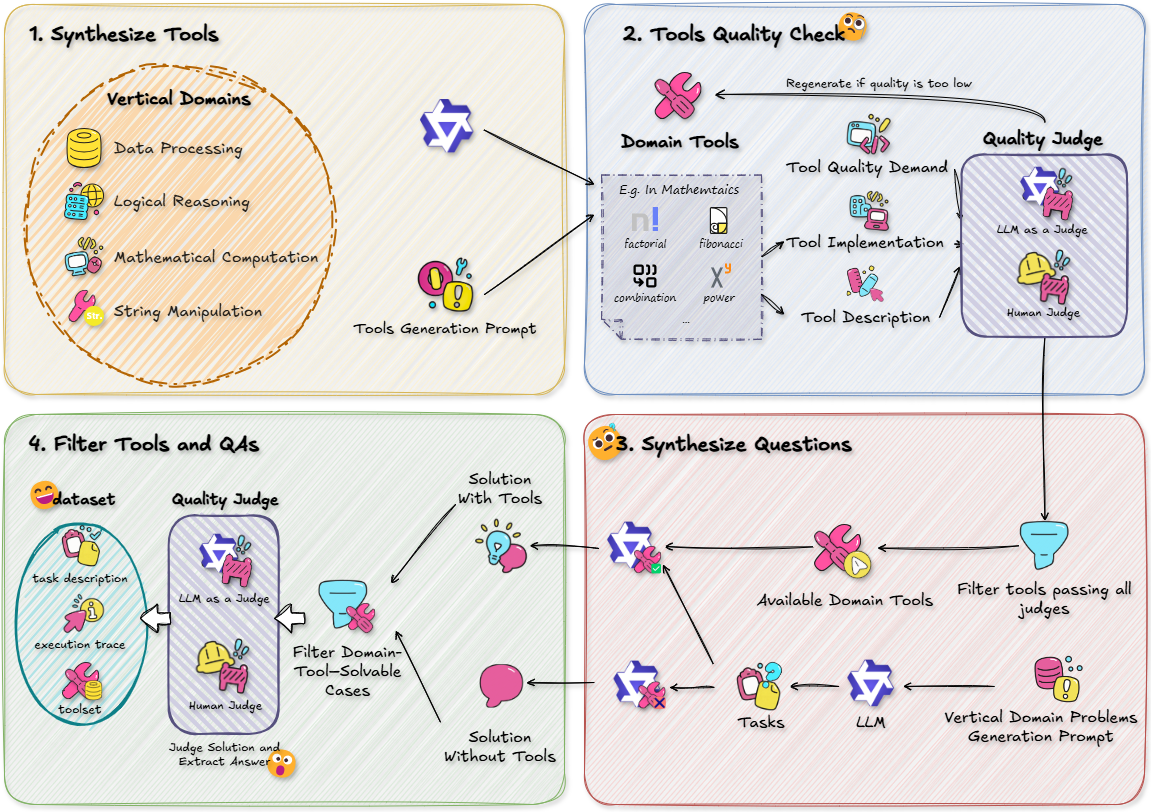}
    \caption{
    Overview of our four-stage dataset construction pipeline. The process synthesizes domain tools, validates tool quality, generates tool-dependent tasks, and filters consistent and high-quality instances to form the final dataset. See Section~\ref{sec:dataset_construction} for details.
    }
    \label{fig:dataset_curation_pipeline}
\end{figure*}

\subsection{Execute-Summarize Design}
We formulate workflow construction as a two-stage process, consisting of an execution phase followed by a summarization phase. This explicit separation contrasts with one-stage approaches such as ReAct or Plan-and-Execute, where reasoning, tool use, and workflow generation are entangled within a single process. In one-stage methods, the model must directly produce a structured workflow while simultaneously solving the task and satisfying representational constraints. Such coupling can introduce unnecessary interference between problem-solving and structural reasoning.

\subsubsection{Execution and Summarization}
\paragraph{Execution Phase.}
The execution phase focuses exclusively on solving the target task by interacting with domain-specific tools. The primary goal is functional correctness, not adherence to workflow representations. During this phase, the model can freely employ any reasoning strategy and invoke tools as needed. The output is an execution trajectory comprising sequences of tool calls, intermediate states, and final results. From the perspective of business problem solving, the Execution phase emphasizes optimal tool selection and strategy adaptation. Models can explore multiple solution paths, experiment with different combinations of tools, and dynamically adjust strategies based on task characteristics.

\paragraph{Summarization Phase.}
The summarization phase operates on recorded execution trajectories to reconstruct a structured workflow abstraction. Unlike execution, this phase is restricted to graph-related tools and has no access to domain-specific business tools. After execution, the model extracts essential steps, dependencies, and control flow to produce a high-level workflow graph, abstracting away low-level execution details while preserving semantic fidelity. From the perspective of workflow construction, Summarization emphasizes information abstraction and structural representation. The model can aggregate multiple execution trajectories to generate the most representative workflow, optimize the granularity of nodes and edges, and apply diverse summarization strategies to produce concise yet semantically rich workflow graphs.

From a modeling perspective, the two-stage design clarifies roles and tool usage: execution leverages the full set of domain tools, while summarization is limited to workflow construction primitives. In contrast, one-stage approaches mix all tools and responsibilities within a single reasoning process. This separation enhances robust task execution and enables more faithful workflow generation.

\subsubsection{Benefits of Decoupling Execution and Summarization}
Decoupling execution from summarization reduces cognitive burden and enhances flexibility. Each stage can adopt independent optimization goals: execution prioritizes exploration and robustness, while summarization focuses on abstraction and structure induction. The framework naturally supports many-to-one mappings, where multiple execution attempts are consolidated into a single workflow, mirroring human trial-and-error learning in which generalized procedures are abstracted from repeated concrete experiences. By grounding summarization in verified execution traces rather than speculative reasoning, the framework reflects how stable strategies emerge from experiential feedback.

Overall, the execute–summarize paradigm provides a robust mechanism for transforming free-form LLM reasoning into structured, executable artifacts. This grounding improves workflow reliability, modularity, and reusability, and enables execution trajectories to support diverse downstream structured representations beyond workflows. Further theoretical framework and extended discussion on cross-domain generalization are provided in Appendix~\ref{sec:theoretical-framework}.

\subsection{FlowBench Overview and Evaluation}
\label{sec:flowbench_summary}

To systematically evaluate FlowMind’s ability, we introduce FlowBench—a synthetic benchmark for structured, multi-step workflow generation. FlowBench consists of three core components: tools, tasks, and execution traces. Tools are curated domain-specific functions, each featuring clearly defined parameters and semantics, all accessible solely through natural language documentation. Tasks are described in natural language and demand coordinated use of multiple tools, presenting non-trivial procedural reasoning challenges that test FlowMind’s problem-solving capabilities. For each task, high-capacity LLMs generate verified execution traces, capturing tool calls, intermediate outputs, and reasoning processes to provide a grounded reference for FlowMind’s workflow induction. The dataset construction pipeline is illustrated in Figure~\ref{fig:dataset_curation_pipeline}, with detailed guidelines and in-depth analysis of the construction process outlined in Appendix~\ref{sec:dataset_construction}. Further details on dataset composition are available in Appendix~\ref{sec:dataset_details}.

We adopt a black-box evaluation paradigm on FlowBench to assess workflow generation. In this framework, model-generated workflows are executed on diverse evaluation cases, and correctness is determined solely by the match between workflow outputs and golden answers, emphasizing functional correctness over structural similarity. White-box analysis is retained as a complementary diagnostic tool, allowing inspection of control-flow structures and tool orchestration (Appendix~\ref{sec:dataset_evaluation}). Together, FlowBench and the black-box evaluation methodology provide a principled, reproducible framework for benchmarking workflow-generation models, capturing both cross-domain generalization and procedural reasoning capabilities.

\section{Experiment}
\begin{table*}[ht!]
\centering
\small
\setlength{\tabcolsep}{6pt}
\begin{tabular}{llcccccc}
\toprule
\textbf{Model} & \textbf{Mode} 
& \textbf{CRate} 
& \textbf{TRate} 
& \textbf{Exec. Compl.} 
& \textbf{Graph Valid.} 
& \textbf{Both Succ.} \\
\midrule
\multirow{4}{*}{Qwen3-8B}
& ReAct            & 5.7  & 6.2  & 63.83 & 34.04 & 25.53 \\
& P\&E             & 17.7 & 21.0 & 27.66 & 73.05 & 23.40 \\
& ES-ReAct         & 18.4 & 23.2 & 64.54 & 73.05 & 51.77 \\
& ES-P\&E          & \textbf{21.3} & \textbf{25.9} & \textbf{88.65} & \textbf{87.23} & \textbf{78.72} \\
\midrule
\multirow{4}{*}{Qwen3-14B}
& ReAct            & 14.9 & 16.0 & 80.85 & 34.04 & 31.21 \\
& P\&E             & 20.6 & 26.5 & 31.21 & 75.18 & 30.50 \\
& ES-ReAct         & 21.3 & 26.7 & 74.47 & 92.20 & 71.63 \\
& ES-P\&E          & \textbf{24.1} & \textbf{27.5} & \textbf{89.36} & \textbf{94.33} & \textbf{85.11} \\
\midrule
\multirow{4}{*}{Qwen3-32B}
& ReAct            & 19.3 & 21.8 & 70.00 & 40.00 & 38.57 \\
& P\&E             & 19.1 & 25.1 & 26.24 & 46.81 & 21.28 \\
& ES-ReAct         & \textbf{21.3} & 24.1 & 73.05 & 73.76 & 57.45 \\
& ES-P\&E          & 19.1 & \textbf{24.9} & \textbf{89.36} & \textbf{87.23} & \textbf{79.43} \\
\midrule
\multirow{4}{*}{GPT-4.1}
& ReAct            & 21.3 & 23.3 & 74.26 & 84.89 & 71.94 \\
& P\&E             & 24.1 & 30.4 & 95.75 & \textbf{97.16} & \textbf{95.04} \\
& ES-ReAct         & 26.2 & 30.4 & \textbf{97.87} & 85.82 & 83.69 \\
& ES-P\&E          & \textbf{29.8} & \textbf{33.3} & 96.45 & 92.91 & 90.78 \\
\midrule
\multirow{4}{*}{GPT-5}
& ReAct            & 26.2 & 30.0 & 68.79 & 66.67 & 54.61 \\
& P\&E             & 32.6 & 35.7 & 43.26 & 76.60 & 43.26 \\
& ES-ReAct         & 23.4 & 28.7 & 78.72 & \textbf{94.33} & 73.76 \\
& ES-P\&E          & \textbf{35.5} & \textbf{36.9} & \textbf{100.00} & 80.85 & \textbf{80.85} \\
\bottomrule
\end{tabular}
\caption{
Main results on FlowBench 141 cases (694 tests).
CRate and TRate denote case-level and test-level pass rates.
Exec. Compl. measures execution completeness,
Graph Valid. indicates the validity of the generated workflow graph,
and Both Succ. reports the fraction of instances where execution and graph generation are both successful.
}
\label{tab:main_results}
\end{table*}

\subsection{Experimental Setup}
We compare four workflow generation paradigms, each representing a different strategy for reasoning and tool orchestration. \textbf{ReAct} interleaves step-by-step reasoning with tool invocation; \textbf{Plan-and-Execute (P\&E)} first generates a complete plan and then executes each step sequentially. 

We further evaluate two variants of our proposed \textbf{Execute-Summarize (ES) framework}: \textbf{ES-ReAct}, which performs task execution using ReAct and subsequently summarizes the execution trace into a reusable workflow graph, and \textbf{ES-P\&E}, which adopts Plan-and-Execute as the execution and summarization strategy.

\subsection{Main Results}
\label{sec:main_results}

Table~\ref{tab:main_results} summarizes the performance of different workflow generation paradigms across model scales on FlowBench.

\paragraph{Overall performance.}
Across all evaluated models, the Execute--Summarize (ES) framework consistently outperforms both ReAct and Plan-and-Execute (P\&E) in terms of case-level and test-level pass rates. In particular, ES-P\&E achieves the strongest overall results in nearly all settings, yielding substantial gains in both execution completeness and graph validity. These improvements translate into markedly higher joint success rates, indicating that execution-aware summarization produces workflows that are not only executable but also structurally reliable.

\paragraph{Execution-Summarize trade-off.}
ReAct tends to favor execution completeness but frequently produces invalid or inconsistent workflow graphs, limiting its joint success rate. In contrast, P\&E improves graph validity by enforcing an explicit planning stage, but often suffers from fragile execution due to plan--execution mismatches. ES-based methods effectively mitigate this trade-off by grounding workflow graph construction in actual execution traces, leading to a much stronger alignment between execution correctness and structural validity.

\paragraph{Effect of model scale.}
While larger models exhibit higher absolute performance across all paradigms, the relative advantage of the ES framework remains stable from Qwen3-8B to GPT-5. This suggests that execution-aware summarization is complementary to model scaling, providing consistent benefits even for strong frontier models.

\paragraph{Variant analysis and method selection.}
In addition to the main paradigms in Table~\ref{tab:main_results}, we evaluate multiple ES-based variants with different execution and graph construction strategies; detailed results are reported in Appendix~\ref{app:es_detailed_results}. While some variants perform well in specific settings, ES-P\&E demonstrates the most consistent and robust performance across model scales and metrics. We therefore adopt ES-P\&E as the default instantiation of FlowMind.

\paragraph{Supporting analyses.}
The conclusions above are supported by extensive diagnostic and ablation analyses in the appendix, including execution--summarization interactions (Appendix~\ref{app:execution_summarization_relationship}), the effect of API-level JSON output constraints (Appendix~\ref{app:json_constraint_ablation}), and workflow graph quality and execution alignment (Appendix~\ref{app:graph_quality_ablation}).
Together, these analyses provide causal evidence for the robustness and design choices of the ES framework.

\subsection{Cognitive Burden Analysis}
\label{sec:cognitive_burden_analysis}

Our task requires models to jointly execute business logic and induce a faithful workflow graph. When these objectives are handled within a single reasoning stage, execution, error handling, abstraction, and tool selection become tightly interleaved, imposing excessive cognitive burden. This coupling leads to systematic failures, including objective interference and entangled tool usage. We show that explicit stage separation alleviates these issues at a structural level.

\subsubsection{Cognitive Burden I: Objective Forgetting}
\label{sec:cognitive_burden_forgetting}

In the vanilla ReAct setting, business execution tools and graph construction tools are exposed simultaneously throughout the interaction.
This design implicitly assumes that the model can retain the intent to construct a workflow graph while navigating long and error-prone execution traces.

In practice, this assumption proves brittle.
We observe that a majority of failures correspond to executions that are largely correct but terminate without producing any workflow graph.
Such omissions account for 66.5\% (133/200) of all failed cases.
This pattern indicates a failure of intent retention rather than insufficient graph construction capability.

To isolate the effect of forgetting from execution difficulty, we introduce an \textbf{Enhanced ReAct} variant.
This setting preserves the single-stage structure and full tool exposure, but introduces an explicit intervention:
if execution completes without graph construction, the model is manually prompted to generate the workflow afterward.
This post hoc enforcement isolates forgetting while keeping all other conditions unchanged.

While Enhanced ReAct reduces omission errors, it does not eliminate cognitive burden. With execution and graph-construction tools still simultaneously available, the model must resolve tool-selection ambiguity after execution, causing spurious actions and unstable graphs.

In contrast, ES-ReAct enforces a strict Execute--Summarize separation: once execution ends, only graph-construction tools are available, eliminating intent ambiguity and letting the model focus on translating the completed trace into a workflow graph.

\begin{figure}[ht!]
    \centering
    \includegraphics[width=\linewidth]{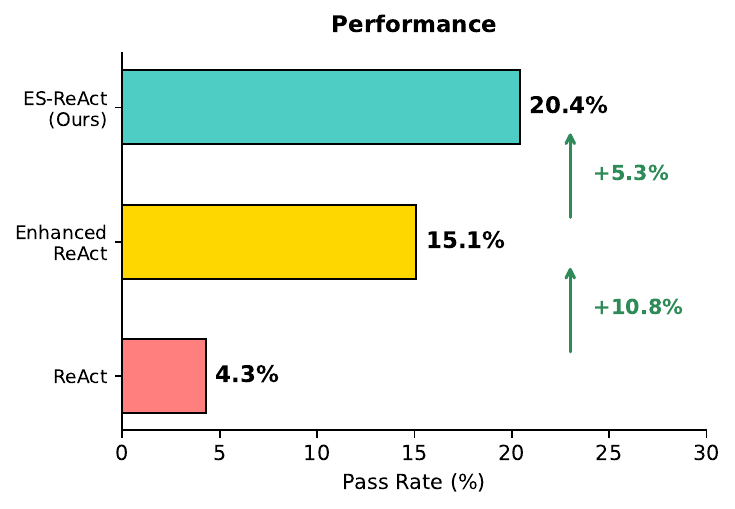}
    \caption{Pass rates of ReAct variants on Qwen3-8B: vanilla ReAct (4.3\%), Enhanced ReAct (15.1\%), and ES-ReAct (20.4\%).}
    \label{fig:react_performance}
\end{figure}

Figure~\ref{fig:react_performance} shows pass rates: vanilla ReAct 4.3\%, Enhanced ReAct 15.1\%, and ES-ReAct 20.4\%, yielding absolute gains of +5.3\% over Enhanced ReAct and +10.8\% over vanilla ReAct.

\subsubsection{Cognitive Burden II: Objective Interference}
\label{sec:cognitive_burden_interference}

Beyond forgetting, cognitive burden also appears as objective interference, where preserving one goal harms another. This is clear in Plan-and-Execute (P\&E), which commits early to a global workflow.

As Table~\ref{tab:main_results} shows, P\&E achieves higher \textit{Graph Valid.} than ReAct, indicating that explicit planning aids structural consistency. However, execution completeness suffers, with large gaps between graph validity and successful execution across all models.

Cognitively, P\&E redistributes rather than removes multi-objective burden. Early plan anchoring prioritizes structure, but execution still requires reacting to feedback, recovering from errors, and aligning a static plan with a dynamic state. This tension leads to premature termination, skipped steps, and incomplete tool use, limiting joint success.

\subsubsection{Cognitive Burden III: Tool-level Interference}
\label{sec:cognitive_burden_tool_interference}

Objective interference is further exacerbated at the interface level by unrestricted access to heterogeneous tools.
We partition tools into two disjoint sets:
business execution tools (B), which advance task progress through concrete operations (e.g., API calls, database queries, or file manipulations), and
graph construction tools (G), which externalize structural abstractions of the execution trace (e.g., creating workflow nodes and edges).
Under ReAct and Plan-and-Execute (P\&E), both tool sets are exposed concurrently without enforced phase boundaries, allowing entangled tool invocation sequences to emerge.

We refer to this phenomenon as \textbf{Mixed Tool Invocation (MTI)}, where tools from B and G are invoked within the same reasoning phase.
Table~\ref{tab:mti_rate} summarizes MTI frequency on Qwen3-8B.
MTI is rare under ReAct but occurs substantially more often in P\&E, reflecting the increased structural pressure induced by early planning and graph commitment.

\begin{table}[ht]
\centering
\small
\setlength{\tabcolsep}{6pt}
\begin{tabular}{lcccc}
\toprule
\textbf{Strategy} & \textbf{Total} & \textbf{Clean} & \textbf{MTI} & \textbf{MTI Rate} \\
\midrule
ReAct            & 200 & 194 & 6  & 3.0\% \\
Plan-and-Execute & 200 & 182 & 18 & 9.0\% \\
\bottomrule
\end{tabular}
\caption{Overall MTI frequency across strategies on Qwen3-8B.}
\label{tab:mti_rate}
\end{table}

Qualitative inspection reveals three recurring MTI manifestations as temporal misalignments between B and G:
\emph{interleaved execution}, where B and G are alternated within a single phase; \emph{interrupted execution}, where the model switches from B to G before execution completes; and \emph{front-loaded graph construction}, where G is committed early and B is subsequently constrained by an incomplete structure. Despite their surface differences, all three prevent the final workflow graph from faithfully reflecting a completed execution trace.

\begin{figure}[ht!]
    \centering
    \includegraphics[width=\linewidth]{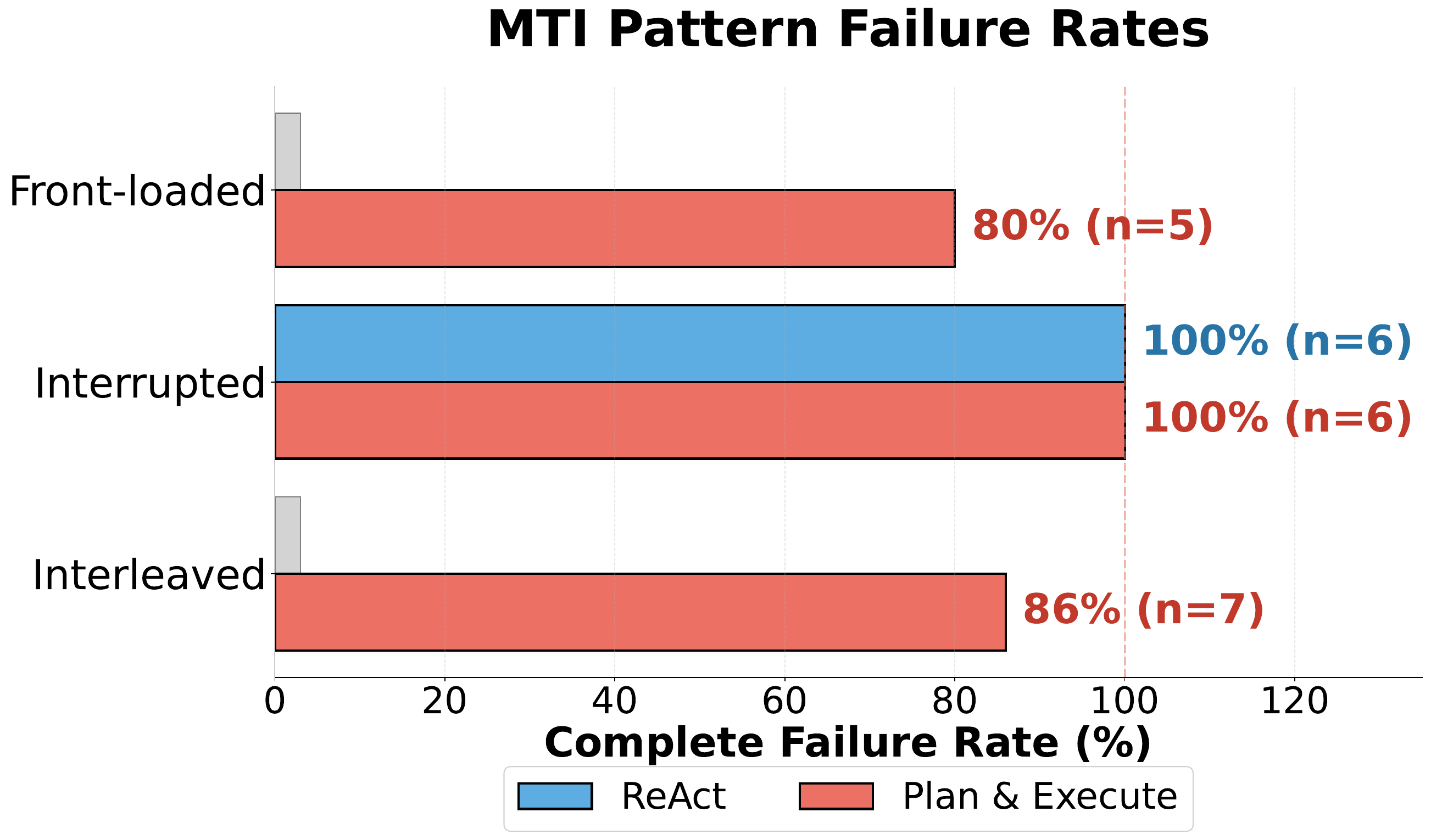}
    \caption{Failure rates on Qwen3-8B across different MTI patterns: interleaved execution and graph construction, interrupted execution, and front-loaded graph construction. Interrupted execution causes 100\% failure for both ReAct and P\&E, highlighting severe tool-level interference.}
    \label{fig:mti_failure_breakdown}
\end{figure}

Figure~\ref{fig:mti_failure_breakdown} shows that such misalignments are strongly correlated with failure.
Interrupted execution results in 100\% failure under both ReAct and P\&E, highlighting tool-level interference as a concrete mechanism linking cognitive burden to degraded execution completeness and graph fidelity.

\subsubsection{Why ES Alleviates Cognitive Burden}
\label{sec:why_es_works}

The Execute--Summarize (ES) framework alleviates cognitive burden through explicit stage separation.
By deferring workflow induction to a dedicated summarize stage, ES prevents objective forgetting and ensures that graph construction operates on completed execution traces.
By decoupling execution from abstraction, ES avoids objective interference, allowing each stage to optimize a single, well-defined goal.
Finally, strict tool partitioning between execution tools (B) and graph construction tools (G) eliminates mixed tool invocation, removing tool-level interference.

\subsection{Impact of Execution Outcomes on Summarization Quality}
\label{sec:execution_impact}

A key advantage of the Execute--Summarize (ES) framework is that it exposes execution as an explicit intermediate signal, enabling direct analysis of how execution outcomes and trajectory quality affect summarization and final performance—an analysis that is difficult in single-stage baselines where execution and graph construction are entangled.

\begin{figure}[t]
    \centering
    \includegraphics[width=\linewidth]{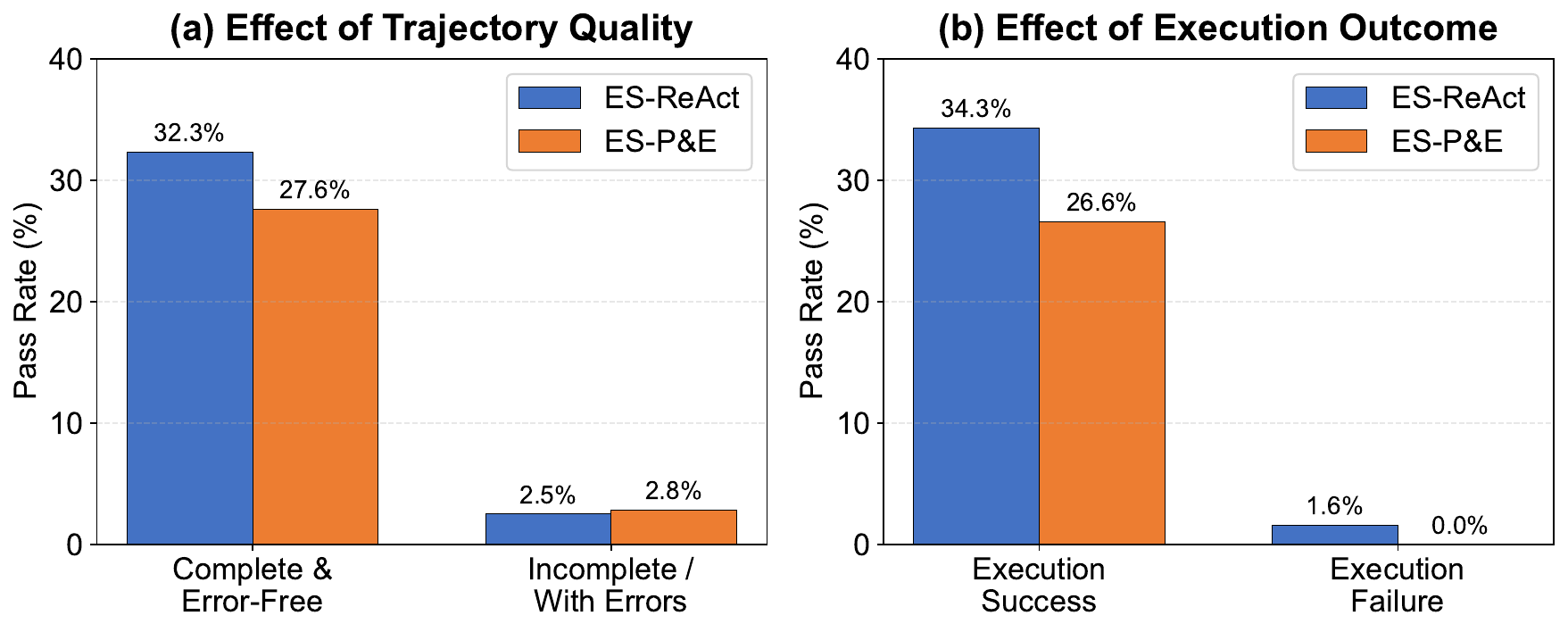}
    \caption{Impact of execution outcomes and trajectory quality on summarization performance on Qwen3-8B. The left panel shows pass rates conditioned on execution success versus failure, while the right panel compares complete and error-free trajectories with incomplete or erroneous ones, for both ES-ReAct and ES-P\&E.}
    \label{fig:execution_impact}
\end{figure}

Figure~\ref{fig:execution_impact} analyzes execution quality along two dimensions. First, conditioning on execution success shows that successful execution is a prerequisite for meaningful performance:
when execution succeeds, summarized workflows achieve non-trivial pass rates (26\%--34\%); when execution fails, pass rates collapse to near zero (0--1.6\%), even if a workflow graph is produced. This indicates that summarization does not compensate for execution failure, but faithfully reflects it.

Second, trajectory integrity further influences outcomes. Complete, error-free traces consistently yield much higher pass rates (above 27\%), whereas incomplete or error-containing trajectories almost always fail (0--3\%). Missing steps and execution errors are thus propagated into the abstracted workflow rather than corrected.

Overall, these results show that ES performance is fundamentally bounded by execution quality: effective summarization depends on successful, complete, and error-free execution traces, confirming that ES gains arise from grounding abstraction in high-fidelity execution signals rather than stronger graph construction alone.

\subsection{Insights from Efficiency and Performance Analysis}
\label{sec:core_insights}
Assessing both practical performance and computational efficiency is essential for validating the effectiveness of the Execute--Summarize (ES) framework in complex workflow generation. Our analysis yields coherent findings explaining why ES outperforms single-stage reasoning paradigms.

A central result is that ES delivers substantial token efficiency while improving task performance. ES variants reduce total output token consumption by 34.0\% to 72.2\% relative to ReAct and P\&E, as redundancy in the execution stage is curtailed—offsetting the modest summarization stage overhead. ES also overcomes performance bottlenecks of single-stage multi-objective reasoning (Appendix~\ref{app:efficiency}).

Deeper diagnostics show ES improves efficiency even when conditioned on outcome: it consumes fewer tokens in successful cases (reflecting concise, goal-aligned trajectories) and mitigates token overuse in failed cases (where single-stage methods waste resources on unproductive reasoning). This stems from structural misalignments in baselines: ReAct suffers intent drift over long trajectories (neglecting graph construction post-business logic), while P\&E over-abstracts prematurely (bypassing necessary tool executions). These flaws reflect the challenge of jointly optimizing execution fidelity and structured output in a single stage (Appendix~\ref{app:performance_gap_diagnosis}).

By separating execution from summarization, ES grounds workflow graphs in verified traces while avoiding single-stage overload. This design jointly improves performance and efficiency, establishing ES as a robust and scalable framework for workflow generation.

\section{Conclusion}
We propose FlowMind, a framework that decouples task execution from structured workflow generation for LLMs. By first completing tasks via Execute phase and then distilling execution traces into coherent workflows by Summarization phase. Our approach enhances the correctness, and modularity of LLM-generated workflows. Empirical results on FlowBench demonstrate that this design substantially improves workflow stability and quality compared to single-stage reasoning paradigms like ReAct and Plan-and-Execute.

\clearpage

\section*{Limitations}
The framework emphasizes concise workflow abstractions for reuse, which entails a degree of compression over execution traces. In tasks with long horizons or subtle intermediate dependencies, some fine-grained behaviors or rare decision points may be less explicitly reflected in the summarized workflow.

\section*{Ethical Considerations}
\label{sec:ethical_considerations}
This work aims to support the responsible study of structured workflow generation systems grounded in large language model (LLM) reasoning. The proposed FlowBench benchmark is constructed entirely from synthetic data, which has been carefully validated to exclude any personal, sensitive, or proprietary information. The FlowMind framework and FlowBench benchmark are intended for research and evaluation purposes. While LLM-based workflow generation shows potential for enhancing automation and interpretability, real-world deployment may introduce risks related to reasoning reliability and error propagation. To mitigate such concerns, our evaluation emphasizes controlled settings and highlights the importance of safeguards such as human oversight and workflow validation.

\section*{Acknowledgments}
This work was supported by National Key Laboratory of Data Space Technology and System.

\bibliography{custom}

\clearpage
\twocolumn[\DoToC]
\clearpage

\appendix

\section{Dataset Details}
\label{sec:dataset_details}

\begin{figure*}[ht]
\centering
\includegraphics[width=\linewidth]{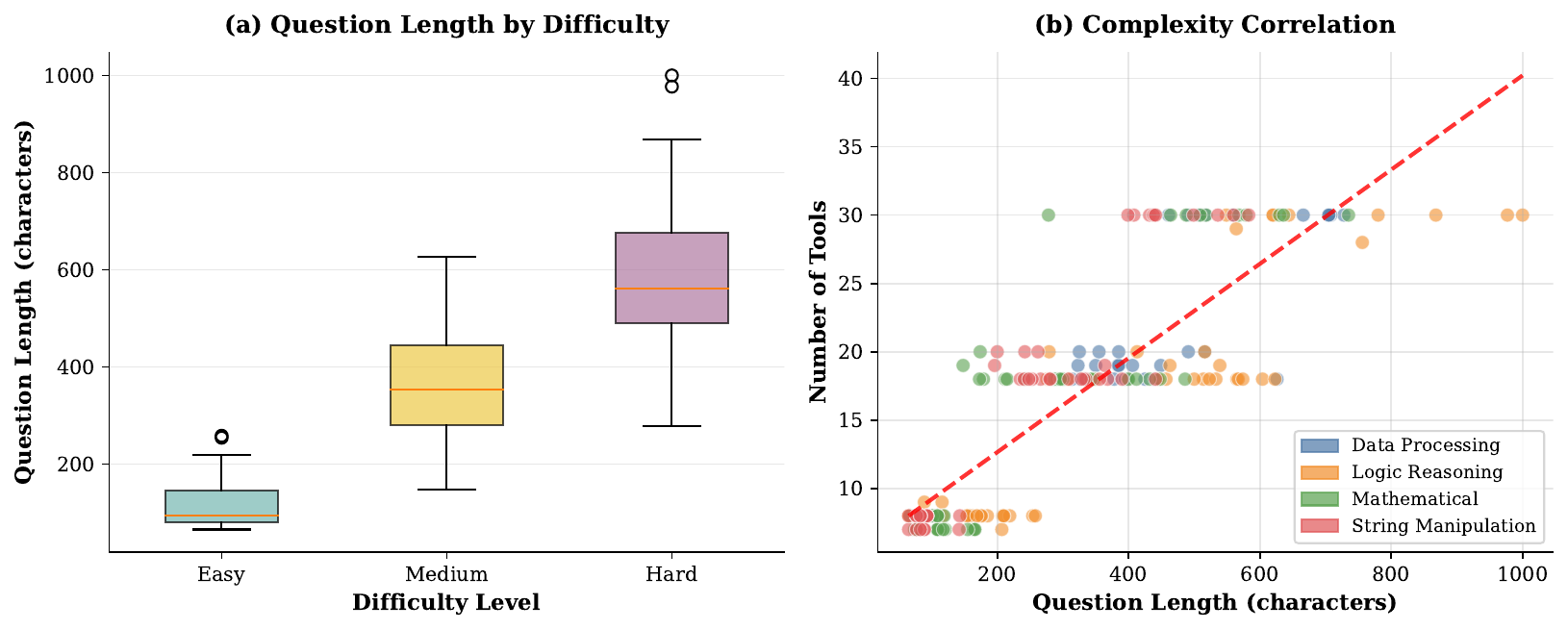}
\caption{Task complexity analysis in FlowBench. (a) Distribution of task description length across three difficulty levels (Easy, Medium, Hard). (b) Correlation between task length and the number of available tools, stratified by four task categories (Data Processing, Logic Reasoning, Mathematical, String Manipulation).}
\label{fig:task_complexity}
\end{figure*}

This section describes the composition, difficulty stratification, and complexity characteristics of the FlowBench dataset, providing a quantitative overview of its structural properties.

\paragraph{Dataset Composition and Statistics.}
\begin{table}[ht!]
\centering
\small
\setlength{\tabcolsep}{4pt}
\begin{tabular}{lcccc}
\hline
Category & Inst. & Tests & Ex. & Tools \\
\hline
Mathematical Computation & 43 & 206 & 103 & 88 \\
Data Processing          & 41 & 203 & 90  & 97 \\
String Manipulation      & 30 & 150 & 81  & 114 \\
Logical Reasoning        & 27 & 135 & 69  & 59 \\
\hline
Total                    & 141 & 694 & 343 & 358 \\
\hline
\end{tabular}
\caption{Dataset statistics across four categories.
``Inst.'' denotes problem instances, ``Tests'' held-out evaluation cases, ``Ex.'' in-context examples, and ``Tools'' unique tools.}
\label{tab:dataset_statistics}
\end{table}

\begin{figure}[t]
    \centering
    \includegraphics[width=0.65\linewidth]{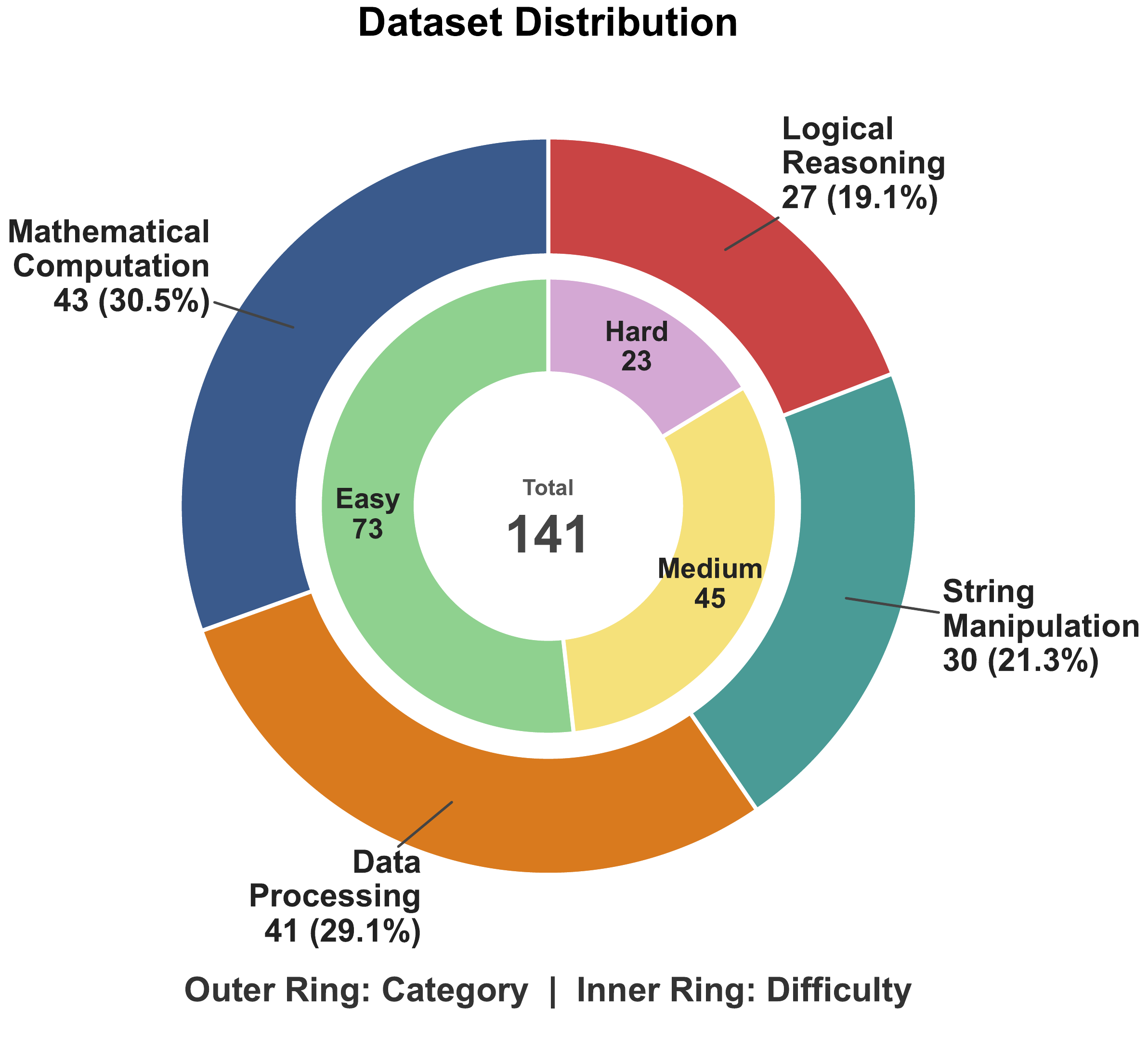}
    \caption{
    Dataset distribution of FlowBench.
    The outer ring shows the category breakdown, including Mathematical Computation (43), Data Processing (41), String Manipulation (30), and Logical Reasoning (27).
    The inner ring indicates task difficulty levels, consisting of Easy (73), Medium (45), and Hard (23).
    }
    \label{fig:flowbench_distribution}
\end{figure}

FlowBench contains \textbf{141 problem instances} spanning four semantic categories: data processing, logical reasoning, mathematical computation, and string manipulation. The number of instances varies across categories, as summarized in Table~\ref{tab:dataset_statistics} and Figure~\ref{fig:flowbench_distribution}

Each problem instance additionally includes a small number of input--output examples for in-context learning and multiple held-out test cases used exclusively for evaluation.

\paragraph{Difficulty Stratification.}
To control task complexity, we stratify the dataset into three difficulty levels based on two orthogonal factors:
(1) the size of the available tool set, and
(2) the required control-flow complexity.
Table~\ref{tab:difficulty_distribution} and Figure~\ref{fig:flowbench_distribution} summarizes the difficulty distribution.

\begin{table}[ht!]
\centering
\small
\setlength{\tabcolsep}{3pt}
\begin{tabular}{lccc}
\hline
Difficulty & Instances & Tools per Instance & Control Flow \\
\hline
Easy   & 73 (51.8\%) & 7--9 (avg.\ 7.7)    & L \\
Medium & 45 (31.9\%) & 18--20 (avg.\ 18.4) & B / I \\
Hard   & 23 (16.3\%) & 28--30 (avg.\ 29.9) & B + I \\
\hline
\end{tabular}
\caption{Difficulty distribution in FlowBench.
Control flow types: L = Linear; B = Branching; I = Iteration.}
\label{tab:difficulty_distribution}
\end{table}

Easy tasks require linear workflows with 2--4 tool invocations. Medium tasks involve either branching or iteration, but not both. Hard tasks combine branching and iteration within the same workflow and expose the model to a large candidate tool set. Difficulty levels are distributed across categories, enabling controlled analysis of performance as task complexity increases.

\paragraph{Task Complexity Analysis.}
Beyond discrete difficulty labels, we further analyze task complexity using continuous structural indicators. Figure~\ref{fig:task_complexity} visualizes the relationship between difficulty levels, task length, and tool-set size.

Several insights emerge from this analysis. First, as shown in the left panel, task descriptions become systematically longer as difficulty increases, reflecting the need to specify richer constraints, conditional logic, and intermediate dependencies. This trend indicates that higher difficulty levels introduce not only larger action spaces but also more complex semantic requirements.

Second, the right panel reveals a positive correlation between task length and the number of available tools. This correlation is consistent across all four categories, suggesting that increased tool-set size naturally induces more elaborate problem formulations. Importantly, this relationship is not category-specific, indicating that FlowBench controls for superficial domain effects while scaling structural complexity.

Taken together, these results demonstrate that FlowBench difficulty levels correspond to meaningful and measurable increases in both linguistic and operational complexity. This supports the use of FlowBench as a controlled testbed for studying how models scale their planning and tool-selection behavior under increasing workflow complexity.

\section{Details of FlowBench Construction}
\label{sec:dataset_construction}

This section presents the design philosophy and construction pipeline of \textsc{FlowBench}.
\textsc{FlowBench} is designed to evaluate models that must plan, orchestrate, and execute structured workflows under realistic tool-use constraints, rather than solving isolated tool calls or single-step reasoning problems.

As illustrated in Fig.~\ref{fig:dataset_curation_pipeline}, \textsc{FlowBench} is constructed through a multi-stage curation pipeline.
Starting from domain and tool synthesis, we progressively generate task specifications and executable workflows with explicit dependency structures and deterministic execution semantics.
We next describe the dataset curation principles, domain and tool construction, and the end-to-end task and workflow synthesis process in detail.

\subsection{FlowBench Curation and Design Principles}
\label{sec:dataset_curation}

\paragraph{Overview.}
FlowBench is a fully synthetic benchmark constructed to study multi-step, tool-mediated reasoning.
All tasks, tools, execution traces, and evaluation instances are generated by large language models and validated through execution.
No real user data is collected at any stage.
All domain content is restricted to abstract, non-personal, and non-sensitive knowledge, ensuring that the dataset does not contain personally identifying information (PII).

The curation of FlowBench is guided by three core principles:
\emph{explicit procedural structure}, \emph{tool-mediated interaction}, and \emph{execution-grounded evaluability}.
These principles distinguish FlowBench from prior benchmarks that focus on single-step function calling or purely textual reasoning.

\paragraph{Explicit procedural structure.}
Each task in FlowBench is constructed such that a correct solution necessarily involves multiple dependent steps.
No task can be solved by a single atomic operation.
Instead, models must construct workflows with intermediate states, control flow, and non-trivial tool composition.
This ensures that benchmark performance reflects genuine planning and orchestration capability rather than surface-level pattern matching.

\paragraph{Tool-mediated interaction.}
Interaction with the environment is exclusively mediated through tools.
Each tool is exposed only via natural-language documentation and a typed parameter interface; executable implementations are hidden from the model.
This setting mirrors realistic agent deployments, where systems must infer tool semantics from documentation rather than relying on source code access.

\paragraph{Execution-based evaluability.}
All tasks admit deterministic execution.
Correctness is defined operationally: a workflow is considered correct if it produces the expected outputs when executed on held-out test cases.
This design enables black-box evaluation based on observable behavior, without relying on latent reasoning traces or surface-form matching.

\subsection{Domain and Tool Construction}
\label{sec:domain_and_tools}

\paragraph{Domain identification.}
Rather than enumerating a fixed taxonomy of application verticals, we identify domains based on functional characteristics that naturally give rise to structured workflows.
These characteristics include information transformation, conditional decision making, iterative processing, aggregation, and symbolic manipulation.
This abstraction-driven selection enables broad domain coverage while avoiding overfitting to domain-specific idiosyncrasies.

\paragraph{Tool specification and curation.}
For each domain, we curate a set of domain-specific tools representing atomic operations.
Each tool is defined by four components:
\begin{itemize}
    \item tool name,
    \item description of its functionality,
    \item typed parameter interface,
    \item description of return values.
\end{itemize}

Tools are intentionally atomic: each tool performs exactly one well-defined operation.
At the same time, semantically related but functionally distinct tools are included (e.g., different filters or comparators), introducing redundancy that requires contextual tool selection.
Tools with ambiguous semantics, excessive overlap, or overly broad functionality are removed through iterative refinement to maintain a clean and interpretable action space.

\subsection{End-to-End Dataset Construction Pipeline}
\label{sec:dataset_pipeline}

Figure~\ref{fig:dataset_curation_pipeline} presents the full construction pipeline of \textsc{FlowBench}, which consists of four tightly coupled stages: \emph{tool synthesis}, \emph{tool quality validation}, \emph{task and workflow synthesis}, and \emph{instance-level filtering}.
Each stage progressively constrains the solution space while preserving diversity and structural complexity.

\paragraph{Stage 1: Domain-driven tool synthesis.}
We begin by prompting a large language model to synthesize candidate tools for each identified domain.
The goal at this stage is to maximize functional coverage and compositional flexibility.
Each synthesized tool corresponds to a single atomic operation and is specified through natural-language documentation and a typed interface, without exposing executable implementations.

\paragraph{Stage 2: Tool quality checking and refinement.}
The synthesized tools are subjected to a quality control process combining LLM-based judgment and manual inspection.
Tools are evaluated for semantic clarity, interface consistency, functional correctness, and non-overlap.
Tools that fail to meet quality requirements are removed or regenerated.
This iterative refinement ensures that the final toolsets form clean and composable action spaces suitable for reliable execution and evaluation.

\paragraph{Stage 3: Task and workflow synthesis.}
Given a validated set of domain tools, we synthesize natural-language tasks that require coordinated use of multiple tools.
Task generation is conditioned on the available tools but does not prescribe explicit solution steps.
We filter out tasks that can be solved via a single tool invocation or without meaningful control flow.
For each retained task, we generate one or more reference execution traces using high-capacity tool-using models.
Each execution trace specifies an ordered sequence of tool calls, including arguments, intermediate outputs, and the final result, forming an explicit executable workflow.

\paragraph{Stage 4: Instance-level filtering and validation.}
To ensure that each instance meaningfully evaluates tool-mediated reasoning, we perform instance-level filtering.
For each task, we compare solutions generated with and without access to tools.
Tasks that can be solved equally well without tool invocation are discarded.
All execution traces are validated through deterministic execution: tool calls must conform to interface specifications, intermediate values must be well-formed, and final outputs must satisfy task requirements.
Only instances that pass all validation checks are included in the final dataset.

\subsection{Task and Workflow Representation}
\label{sec:task_construction}

Each instance in FlowBench is represented as a triplet:
(\textbf{task description}, \textbf{toolset}, \textbf{execution trace}).
During evaluation, models are provided with the task description and tool documentation, but not the reference execution trace.
This separation supports both trace-free evaluation and trace-aware learning paradigms, while ensuring fair comparison across methods.

\subsection{License}

The code associated with this work is released under the Apache License, Version 2.0.
The Apache-2.0 license permits use, reproduction, modification, and distribution, and includes an express patent license from contributors, subject to the license conditions.
The accompanying dataset and documentation are also released under Apache-2.0, unless otherwise stated.

\subsection{Intended Use and Compliance}

The dataset is generated via Claude and Qwen APIs for research and reproducibility.
We do not distribute raw API-generated responses.
Instead, we release prompts, generation scripts and parameters, and evaluation code, allowing others to reproduce the dataset using their own compliant API access.

\subsection{Potential Risks and Limitations}

Despite its controlled design, FlowBench has several inherent limitations.
First, its abstract and synthetic nature may limit direct transferability to real-world tool-use scenarios, where tools often exhibit ambiguous semantics or undocumented constraints.
Second, reliance on LLM-generated tasks and execution traces may propagate biases present in the base models, such as overrepresentation of canonical reasoning patterns.
Finally, deterministic validation, while ensuring consistency, may filter out valid but non-canonical workflows, potentially narrowing the diversity of evaluated solution strategies.

\subsection{Instructions Given to Participants}

All human-in-the-loop verification followed a written instruction sheet describing the task goals, verification procedures, and acceptance criteria.
Annotators were instructed to verify each candidate instance, correct unclear wording while preserving original intent, and reject any item with ambiguity or inconsistency.
Only instances receiving unanimous approval were retained.
The instruction sheet also included an ethics and safety notice prohibiting the introduction or disclosure of personal data.

\subsection{Characteristics of Annotators}

All annotation and data filtering for FlowBench were conducted collaboratively by the authors and large language models, without involving external annotators.
This design ensures strong alignment between annotation criteria and research objectives, while maintaining high consistency through domain expertise and model-assisted efficiency.

\section{Details of FlowBench Evaluation}
\label{sec:dataset_evaluation}

In our setting, a domain query is mapped to a \emph{workflow} as the final output of FlowMind. Therefore, the central evaluation challenge is no longer how to solve the domain problem itself, but how to determine whether a generated workflow is \emph{correct}. Judging the correctness of a workflow---analogous to judging the correctness of a program or a piece of code---is inherently non-trivial. We consider two complementary evaluation paradigms to address this challenge: \emph{white-box} evaluation and \emph{black-box} evaluation. Figures~\ref{fig:white_box_evaluation} and~\ref{fig:black_box_evaluation} illustrate the two designs, respectively.

\subsection{White-Box Evaluation}
\begin{figure*}[!t]
  \centering
  \includegraphics[width=\linewidth]{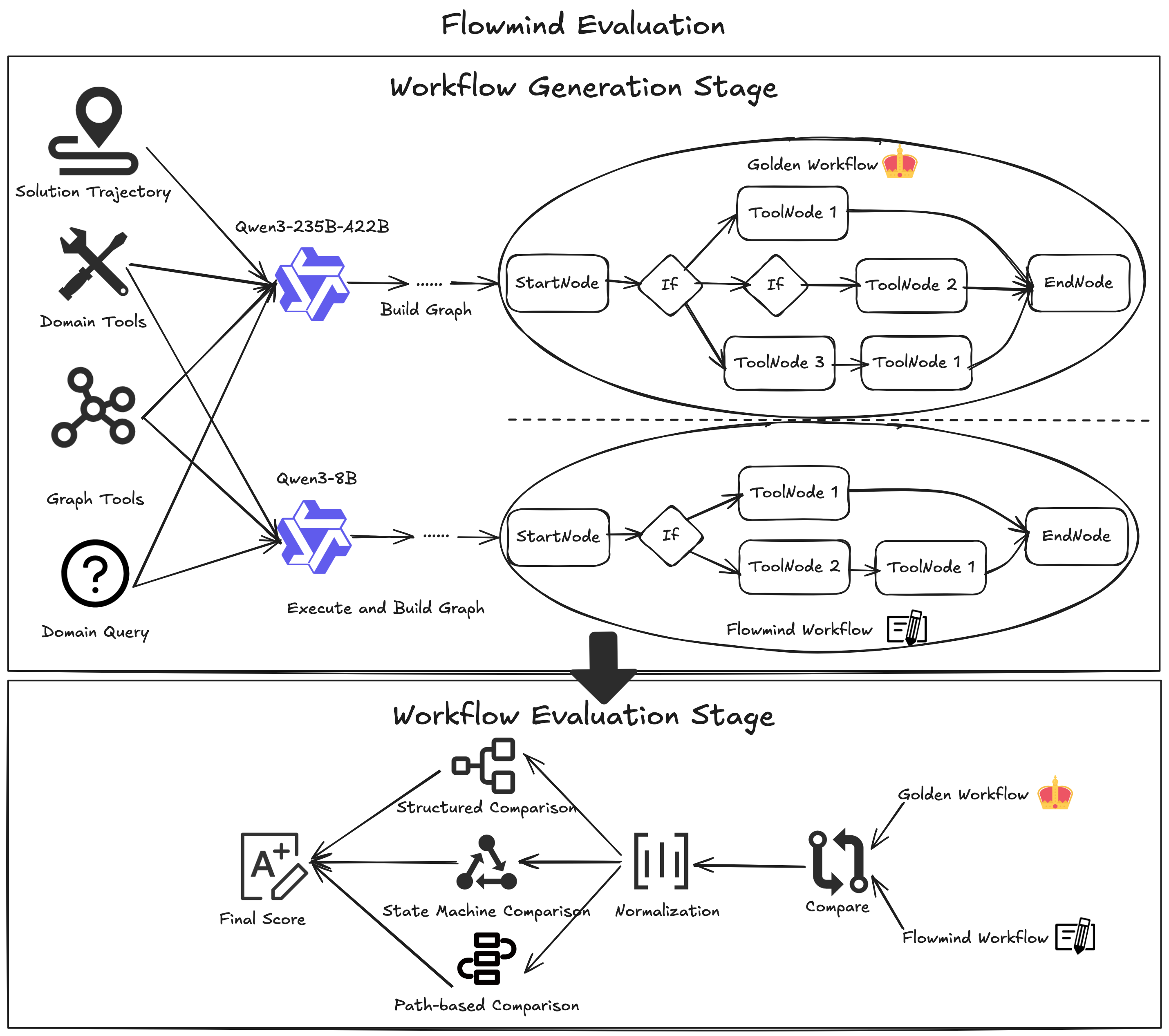}
  \caption{White-box evaluation of FlowMind. The correctness of the FlowMind-generated workflow is statistically determined through multi-dimensional consistency comparison with the Golden Workflow.}
  \label{fig:white_box_evaluation}
\end{figure*}

White-box evaluation assesses workflow generation by directly inspecting the internal structure and execution behavior of a workflow, rather than relying solely on final answers. Under this paradigm, a workflow is considered correct if its control flow, decision structure, and tool usage align with those of an ideal solution.

As illustrated in Figure~\ref{fig:white_box_evaluation}, the evaluation framework consists of two stages: a \emph{workflow generation stage} and a \emph{workflow evaluation stage}. In the generation stage, two workflows are constructed for the same domain query, but under fundamentally different input conditions. The first is a \emph{Golden Workflow}, which serves as the reference. During dataset construction, each domain query is paired with a high-quality solution and a verified final answer. These solutions explicitly describe the correct reasoning steps, decision points, and tool usage required to solve the task. Leveraging this information, we employ a high-capacity LLM to transform the solution and answer into a structured control-flow graph. This process is performed offline and does not involve exploration or tool execution; instead, the model deterministically maps an already-correct reasoning trajectory into a standardized workflow representation consisting of start and end nodes, conditional branches, and tool nodes. As a result, the Golden Workflow represents a faithful structural abstraction of the ideal solution.

In contrast, the \emph{FlowMind Workflow} is generated online by a smaller LLM using only the original domain query and the available tool set, without access to the solution or answer. The model must autonomously plan, make conditional decisions, and invoke tools during execution, while simultaneously constructing its workflow graph. This intentional information asymmetry ensures that the evaluation measures the model’s ability to recover the correct control-flow structure from the problem itself, rather than its ability to reproduce a known solution.

During workflow evaluation, both the Golden Workflow and the FlowMind-generated candidate are first normalized into a canonical representation that abstracts away superficial discrepancies in node identifiers, condition syntax, and graph layout.  
The resulting graphs are rigorously contrasted along three complementary axes.  
\textbf{Structural congruence} quantifies the overlap of nodes, edges, and branching topologies.  
\textbf{Behavioral equivalence} reduces each workflow to a deterministic finite-state machine and verifies that every state–transition function and condition-triggered action coincides.  
\textbf{Path sufficiency} enumerates the complete set of feasible execution traces to confirm that the candidate covers every critical decision path mandated by the Golden Workflow while introducing neither spurious nor redundant traces.

The results of these comparisons are aggregated into a final score that quantifies the degree of structural and behavioral similarity between the generated workflow and the ideal reference. By explicitly evaluating how a workflow is constructed and executed, rather than only what answer it produces, this white-box evaluation provides a fine-grained and interpretable measure of a model’s planning, decision-making, and tool orchestration capabilities.

\subsection{Black-Box Evaluation}

\begin{figure*}[!t]
  \centering
  \includegraphics[width=\linewidth]{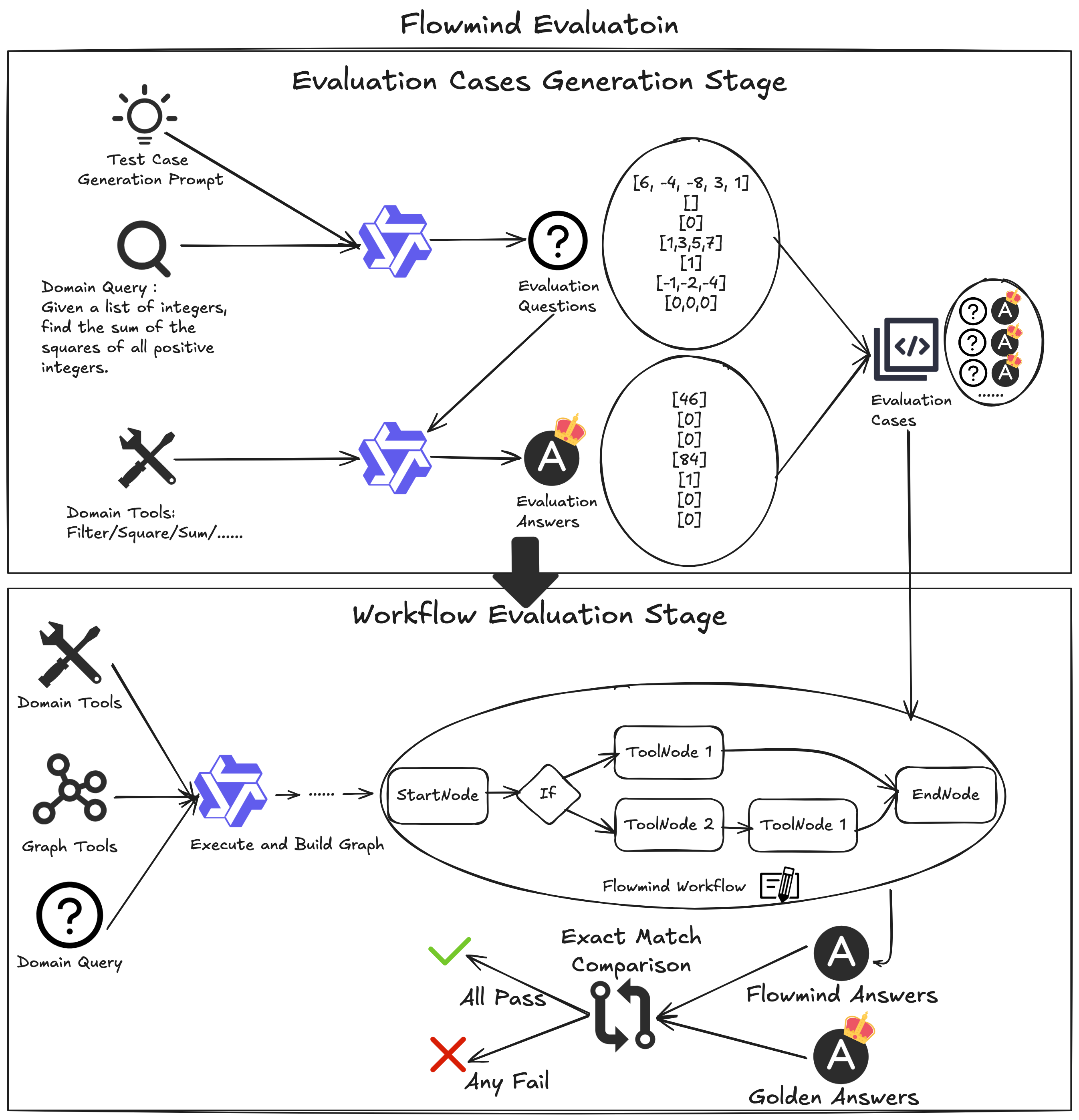}
  \caption{Black-box evaluation of FlowMind. The generated workflow is treated as a black box and judged by its performance on a suite of domain-specific test cases, similar to program evaluation via unit tests.}
  \label{fig:black_box_evaluation}
\end{figure*}

Black-box evaluation provides a practical and outcome-oriented paradigm for assessing workflow generation, aligning closely with how workflow systems are ultimately used in real-world applications. Instead of defining correctness through explicit inspection of internal structures, this paradigm evaluates a workflow purely based on its observable behavior: whether it consistently produces correct outputs when executed on concrete instances of the domain query. From this perspective, a workflow is regarded as a functional artifact whose internal realization is secondary to its end-to-end correctness and reliability.

As illustrated in Figure~\ref{fig:black_box_evaluation}, the black-box evaluation framework is composed of two stages: an \emph{evaluation case generation stage} and a \emph{workflow evaluation stage}. In the evaluation case generation stage, the abstract domain query is instantiated into a diverse set of concrete evaluation cases. Both the generation of concrete inputs (i.e., golden queries) and the computation of corresponding golden answers are performed by high-capacity models with strong reasoning and tool-use capabilities, in conjunction with verified domain tools. These models are reliable to produce correct and consistent evaluation cases. By delegating evaluation case construction to substantially stronger models, we significantly reduce the risk of noisy or incorrect supervision and ensure that the resulting evaluation suite faithfully reflects the intended semantics of the domain query.

In the workflow evaluation stage, the FlowMind-generated workflow is executed on each evaluation case in a fully autonomous manner. The workflow dynamically plans, branches, and invokes domain tools according to its internally constructed logic. Throughout this process, the evaluator remains agnostic to the workflow’s internal control-flow structure and decision-making process, focusing exclusively on the final outputs. For each evaluation case, the workflow’s output is compared against the corresponding golden answer using an exact-match criterion. A workflow is considered correct under black-box evaluation only if it produces correct outputs for all evaluation cases, yielding a strict and unambiguous notion of correctness.

An important advantage of this evaluation paradigm is that it decouples correctness from any particular reasoning trace or workflow topology. Different workflows may employ distinct tool invocation orders, branching strategies, or intermediate representations, yet still correctly solve the task. Black-box evaluation naturally accommodates this diversity by judging workflows solely on their functional behavior. This property is especially desirable in the context of workflow generation, where multiple structurally distinct solutions may be equally valid.

Furthermore, because evaluation correctness is grounded in high-quality, model-verified golden queries and answers, black-box evaluation provides a strong and reliable signal of true task-solving capability. It scales efficiently to large evaluation sets and closely mirrors how workflows are expected to perform in deployment: given arbitrary valid inputs, the workflow must reliably produce correct outputs.

\subsection{Choice of Evaluation Paradigm}

Based on the above analysis, we adopt \emph{black-box evaluation} as the primary evaluation paradigm in this work. 
This decision is driven by both practical considerations and fundamental limitations of workflow-level structural comparison.
We summarize our motivations as follows.

\paragraph{Limitations of workflow-level structural comparison.}
Directly comparing workflows is intrinsically difficult, especially at the semantic level:

\begin{itemize}
    \item \textbf{Semantic alignment is non-trivial.} 
    Even after normalization, workflow comparison requires resolving semantic equivalence between conditions, aligning branching logic, and establishing correspondence between intermediate states.
    Minor syntactic differences may conceal deep semantic equivalence, while superficially similar structures may diverge under corner cases.

    \item \textbf{Opaque tools and APIs break white-box assumptions.}
    Real-world workflows inevitably rely on domain-specific business APIs whose internal implementations are unobservable.
    Under such conditions, it is often impossible to determine whether two different sequences of tool invocations are functionally equivalent.
    Structurally distinct workflows may use different APIs, execution orders, or intermediate representations, yet yield identical outcomes.

    \item \textbf{Workflow equivalence is hard even for humans.}
    Constructing and verifying complex workflows with nested branches and dynamic control flow is cognitively demanding.
    Determining whether two workflows are equivalent across all execution paths is notoriously difficult, even for human experts.
    Expecting large language models to reliably perform such judgments at scale is therefore unrealistic.
\end{itemize}

\paragraph{Advantages of black-box evaluation.}
In contrast, black-box evaluation offers several decisive benefits:

\begin{enumerate}
    \item \textbf{Focus on functional correctness.} 
    Black-box evaluation abstracts away internal implementation details and evaluates workflows purely based on input--output behavior.
    This aligns with real-world business intuition: users care about whether a workflow achieves the desired functionality, not how it internally orchestrates tools.

    \item \textbf{Scalable and practical test generation.}
    Generating high-quality test cases is substantially easier than synthesizing or verifying full workflows.
    Strong large language models, combined with reliable domain tools, can instantiate abstract queries into diverse and representative concrete cases, enabling comprehensive evaluation coverage.

    \item \textbf{Ease of human validation.}
    The evaluation process closely resembles unit testing or programming benchmarks such as LeetCode, where correctness is judged by observable outputs.
    Because tool calls and final results are explicit, human evaluators can readily verify the reasonableness of execution traces and the reliability of golden answers.
\end{enumerate}

\paragraph{Summary.}
Taken together, these considerations make black-box evaluation a more robust, scalable, and practically meaningful choice for assessing workflow generation.
While white-box evaluation remains valuable for diagnostic analysis and interpretability, black-box evaluation provides a clearer and more reliable signal of end-to-end task-solving capability.
Accordingly, we adopt black-box evaluation as our primary evaluation paradigm, and use white-box analysis as a complementary tool for qualitative understanding and error analysis.

\section{Theoretical Framework and Cross-Domain Generalization}
\label{sec:theoretical-framework}

At its core, \textsc{FlowMind} is built on a simple but fundamental principle: 
\emph{behavioral correctness and structural abstraction should be optimized at different stages}.
Rather than committing to a complete workflow before interacting with the environment, FlowMind first focuses on successfully executing the task under real conditions, and only afterward abstracts the observed behavior into a reusable workflow. This separation provides both a formal foundation for reasoning and a natural explanation for the paradigm’s robustness and cross-domain generality.

\subsection{Formal Foundations}

We model a tool-augmented task as a Markov Decision Process (MDP):
\[
\mathcal{T} = (\mathcal{S}, \mathcal{A}, \mathcal{P}, \mathcal{R}, s_0, s^*),
\]
where $\mathcal{S}$ denotes the state space (including observable task progress, latent tool states, and contextual information), $\mathcal{A}$ the action space (tool invocations and intrinsic reasoning steps), $\mathcal{P}$ a stochastic transition function capturing nondeterministic tool or environment responses, and $\mathcal{R}$ a reward function assigning $1$ to task completion and non-positive values to invalid or redundant actions. The task begins at $s_0$ and terminates upon reaching the target state $s^*$.

\paragraph{Execution Trace.}
An execution trace is defined as
\[
\tau = (s_0, a_0, s_1, a_1, \dots, s_T, a_T),
\]
where $s_{t+1} \sim \mathcal{P}(s_t, a_t)$ and $T$ denotes the termination step.

\paragraph{Workflow Abstraction.}
A workflow is a structured control-flow abstraction
\[
\mathcal{W} = (N, E),
\]
where $N$ is a set of semantically coherent nodes, each corresponding to a subset of trace segments, and $E$ encodes control relations such as sequencing, branching, and iteration.

\paragraph{Workflow Fidelity.}
We define fidelity between a workflow $\mathcal{W}$ and a trace $\tau$ as
\[
\mathcal{F}(\mathcal{W}, \tau) \in [0,1],
\]
measuring how well the workflow preserves the causal and sequential dependencies observed in execution.

\subsection{Paradigms as Different Abstraction Timings}

Existing agent paradigms can be unified by a single dimension: \emph{when abstraction is introduced relative to execution}.

\paragraph{ReAct.}
ReAct performs single-stage online reasoning, selecting an action at each timestep via a myopic value estimate:
\[
\begin{aligned}
a_t
&= \arg\max_{a \in \mathcal{A}}
   \mathbb{E}_{s_{t+1} \sim \mathcal{P}(s_t,a)} \\
&\quad \Bigl[
    \mathcal{R}(s_t,a)
    + \gamma V_{\mathrm{ReAct}}(s_{t+1})
\Bigr].
\end{aligned}
\]

Because long-horizon dependencies are not explicitly represented, value estimation errors compound over time, leading to redundant tool calls and unstable trajectories.

\paragraph{Plan-and-Execute.}
Plan-and-execute approaches attempt to generate a complete workflow prior to execution:
\[
\mathcal{W}_{\mathrm{plan}} =
\arg\max_{\mathcal{W} \in \mathcal{W}_{\mathrm{all}}}
\mathbb{P}(\mathrm{Execute}(\mathcal{W}, s_0) \rightarrow s^*).
\]
However, planning is performed without access to true state transitions, particularly latent tool states. As a result, execution often deviates from the planned structure, yielding low workflow fidelity:
\[
\mathcal{F}(\mathcal{W}_{\mathrm{plan}}, \tau_{\mathrm{exec}})
\le 1 - \delta(\mathcal{S}_{\mathrm{latent}}),
\]
where the penalty grows with the complexity of unobserved states.

\paragraph{FlowMind.}
FlowMind adopts a two-stage, trace-grounded process:
\begin{itemize}
    \item \textbf{Execution stage.} An executor interacts with the environment to produce a successful trace
    \[
    \tau^* = \arg\max_{\tau \in \mathcal{T}_{\mathrm{all}}}
    \mathbb{P}(\tau \rightarrow s^*),
    \]
    ensuring correctness through real-time feedback.
    \item \textbf{Summarization stage.} A summarizer abstracts the observed trace into a workflow
    \[
    \mathcal{W}^* =
    \arg\max_{\mathcal{W} \in \mathcal{W}_{\mathrm{all}}}
    \mathcal{F}(\mathcal{W}, \tau^*),
    \]
    yielding high-fidelity structure aligned with actual execution.
\end{itemize}
As task horizon grows, the resulting workflow satisfies
\[
\mathcal{F}(\mathcal{W}^*, \tau^*) = 1 - o(1),
\]
preserving causal detail without sacrificing abstraction.

\subsection{Implications: Robustness, Generalization, and Efficiency}

This execution-first abstraction strategy has several important consequences that extend beyond any specific domain or tool set.

\paragraph{Robustness under Uncertainty.}
Because workflows are synthesized from real execution traces, they naturally encode retries, conditional branches, and recovery behaviors induced by stochastic tools or unforeseen constraints. Unlike pre-specified plans that rely on idealized assumptions, FlowMind internalizes mismatches between expected and actual behavior, making adaptation an intrinsic property rather than an exceptional case. These advantages arise whenever environments exhibit partial observability or nondeterminism, which is common across real-world domains.

\paragraph{Interpretability and Reusability.}
Each workflow step in FlowMind is grounded in execution evidence, establishing a direct correspondence between abstract structure and observed behavior. This improves interpretability and auditability, which is critical in knowledge-intensive or regulated settings. Moreover, the summarization process yields modular abstractions: semantically coherent segments of behavior become reusable components that can be adapted or recombined without reconstructing the entire procedure.

\paragraph{Computational Efficiency.}
FlowMind avoids the combinatorial explosion of exhaustive planning. Instead of solving an exponential search problem over large action spaces, it incurs linear execution cost followed by lightweight abstraction over a compact trace. If $T = |\tau|$ and $K = |\mathcal{W}| \ll T$, the total cost satisfies
\[
\mathrm{Cost}_{\mathrm{FlowMind}} = O(T + K),
\]
making the paradigm suitable for large-scale or long-horizon systems where traditional planning is infeasible.

\subsection{Summary}

By decoupling behavioral correctness from structural abstraction, FlowMind provides a principled framework for transforming unstructured sequential behavior into reliable and reusable workflows. This trace-grounded design explains both its theoretical advantages over existing paradigms and its ability to generalize across domains characterized by uncertainty, interaction, and complex tool use.

\section{Ablation and Diagnostic Analyses}

\subsection{Effect of API-Level JSON Output Constraints}
\label{app:json_constraint_ablation}

To examine a key design choice in our function-calling pipeline, we conduct an ablation study on the effect of enforcing structured JSON output through API-level constraints. Many modern LLM inference frameworks, including OpenAI APIs and vLLM, provide a \texttt{response\_format} option that forces the model to output valid JSON objects. While such constraints are generally assumed to improve reliability for structured generation, our experiments reveal a counterintuitive and severe degradation in function-calling performance when applied to smaller models.

\paragraph{Experimental Setup.}
We evaluate two configurations using the Qwen3-8B model deployed via vLLM:
\begin{itemize}
    \item \textbf{With JSON Constraint(w JSON)}: All LLM calls are enforced with \texttt{response\_format=\{"type": "json\_object"\}}.
    \item \textbf{Without JSON Constraint(w/o JSON)}: No format constraint is applied; structured outputs are recovered using a robust post-processing pipeline.
\end{itemize}

Both configurations are tested under four orchestration strategies: \textsc{ReAct}, \textsc{Plan-and-Execute} (P\&E), \textsc{Execute-Build-Graph} with ReAct-style execution, and \textsc{Execute-Build-Graph} with P\&E-style execution.

\paragraph{Results.}
Table~\ref{tab:json_constraint_ablation} reports the pass rates under the two configurations. Enforcing the JSON constraint results in complete failure across all strategies, yielding a 0\% pass rate. In contrast, removing the constraint leads to dramatic performance recovery: three out of four strategies achieve perfect accuracy on the evaluated test case.

\begin{table}[ht!]
\centering
\small
\setlength{\tabcolsep}{6pt}
\begin{tabular}{lcc}
\toprule
\textbf{Strategy} 
& \textbf{JSON Constrained} 
& \textbf{Unconstrained} \\
\midrule
ReAct    & $\times$ & $\checkmark$ \\
P\&E     & $\times$ & $\times$ \\
ES-ReAct & $\times$ & $\checkmark$ \\
ES-P\&E  & $\times$ & $\checkmark$ \\
\bottomrule
\end{tabular}
\caption{Ablation on API-level JSON output constraints. Strict JSON formatting prevents successful execution across all strategies.}
\label{tab:json_constraint_ablation}
\end{table}

\paragraph{Error Analysis.}
A closer inspection of failure cases under the JSON-constrained setting reveals a consistent error pattern: \emph{``No Graph available for validation.''} In these cases, the model fails to produce any valid function call, resulting in the absence of both execution trajectories and constructed graphs.

Analysis of the raw model outputs indicates that when forced into pure JSON mode, the 7B model frequently generates malformed or incomplete JSON structures. This issue is particularly pronounced when the model attempts to reconcile internal reasoning with structured action specifications. The hard constraint effectively suppresses natural language reasoning, which appears to destabilize the model’s learned function-calling behavior.

In contrast, without the JSON constraint, the model freely interleaves natural language reasoning with embedded JSON-formatted function calls. Our post-processing pipeline—comprising multi-JSON extraction, heuristic validation, and retry mechanisms—successfully recovers valid function calls from these mixed-format outputs.

\paragraph{Discussion.}
This ablation study highlights several important implications for agentic systems built on smaller language models:
\begin{enumerate}
    \item \textbf{Structured Constraints Can Harm Smaller Models.} Although JSON mode is designed to improve output correctness, it may conflict with how smaller models internalize the coupling between reasoning and action. Forcing strict structural compliance can disrupt function-calling altogether.
    \item \textbf{Robust Parsing Is Preferable to Hard Constraints.} Instead of relying on API-level enforcement, we find that flexible generation combined with robust downstream parsing yields substantially better results.
    \item \textbf{Model Capacity Matters.} This failure mode appears particularly severe for 7B-scale models. Larger models (e.g., 70B+) may better tolerate strict format constraints, though this hypothesis warrants further study.
\end{enumerate}

\paragraph{Design Decision.}
Based on these findings, we remove the \texttt{response\_format} constraint from our final system design. Instead, we adopt a multi-stage parsing strategy that (i) attempts standard JSON parsing, (ii) falls back to multi-JSON extraction for mixed-format responses, and (iii) applies corrective prompting with retries when necessary. This design enables strong performance under resource-constrained settings while remaining compatible with larger models that may not require such flexibility.

\paragraph{Takeaway.}
As shown in Table~\ref{tab:json_constraint_ablation}, strict API-level JSON enforcement can catastrophically impair function-calling performance for small models. These results suggest that hard structural constraints are not universally beneficial and may conflict with learned reasoning--action coupling. Accordingly, we adopt flexible generation with robust downstream parsing as the default design choice in all main experiments.

\subsection{Workflow Graph Quality and Execution Alignment}
\label{app:graph_quality_ablation}

To isolate the factors underlying performance differences in tool-augmented workflow generation, we conduct an ablation study on \emph{workflow graph construction quality} and its \emph{alignment with execution outcomes}. We focus on a curated 141-case subset of FlowBench (Section~\ref{sec:flowbench_summary}), where all cases are solvable by a high-capacity baseline model (Claude). This setting controls for task ambiguity and unsolvability, allowing failures to be primarily attributed to execution and graph construction rather than task definition.

\paragraph{Experimental Setup.}
We evaluate Qwen3-8B under four orchestration strategies: \textsc{ReAct}, \textsc{Plan-and-Execute (P\&E)}, and two variants of our proposed \textsc{Execute-Summarize (ES)} framework, namely ES-ReAct and ES-P\&E. For each case, we record execution success, workflow graph correctness with respect to the golden workflow, and the final end-to-end pass rate under black-box evaluation.

\paragraph{Overall Performance.}
Table~\ref{tab:141_overall} summarizes end-to-end pass rates on the 141-case subset.

\begin{table}[ht!]
\centering
\small
\setlength{\tabcolsep}{8pt}
\begin{tabular}{lcc}
\toprule
\textbf{Strategy} & \textbf{Passed / Total} & \textbf{Pass Rate} \\
\midrule
ReAct & 8 / 141 & 5.7\% \\
Plan-and-Execute & 25 / 141 & 17.7\% \\
ES-P\&E & 28 / 141 & 19.9\% \\
ES-ReAct & \textbf{32 / 141} & \textbf{22.7\%} \\
\bottomrule
\end{tabular}
\caption{Pass rates of Qwen3-8B on the FlowBench 141-case subset under different orchestration strategies.}
\label{tab:141_overall}
\end{table}

ES-ReAct outperforms the vanilla ReAct baseline. Decoupling execution from workflow graph construction substantially improves small-model performance on tool-use tasks.

\paragraph{Performance by Task Category.}
Performance broken down by FlowBench task category is reported in Table~\ref{tab:141_by_category}.

\begin{table}[ht!]
\centering
\small
\setlength{\tabcolsep}{2.5pt}
\begin{tabular}{lcccc}
\toprule
\textbf{Category} & \textbf{ReAct} & \textbf{P\&E} & \textbf{ES-ReAct} & \textbf{ES-P\&E} \\
\midrule
Data Proc. (41)    & 7.3\%  & 36.6\% & \textbf{48.8\%} & 46.3\% \\
String Manip. (30)& 13.3\% & 20.0\% & \textbf{23.3\%} & 16.7\% \\
Logic Reason. (27)& 3.7\%  & \textbf{14.8\%} & 7.4\% & 11.1\% \\
Math Comp. (43)   & 0.0\%  & 0.0\%  & \textbf{7.0\%} & 2.3\% \\
\bottomrule
\end{tabular}
\caption{
Category-wise pass rates (\%) on the 141-case FlowBench subset.
 P\&E denotes Plan-and-Execute and ES-P\&E denote its Execute-Summarize variants. Numbers in parentheses indicate the number of tasks per category.
}
\label{tab:141_by_category}
\end{table}

Three consistent patterns emerge. First, \emph{Data Processing} tasks benefit most from the ES framework: ES-ReAct nearly halves the remaining error rate compared to ReAct, indicating strong alignment between execution traces and structured workflow summarization. Second, \emph{Logical Reasoning} tasks favor explicit pre-planning, where Plan-and-Execute outperforms ES variants, suggesting that multi-condition evaluation and branching logic are better handled by global plans. Third, \emph{Mathematical Computation} remains challenging across all strategies, implying that execution strategy optimization alone cannot compensate for limitations in symbolic or arithmetic reasoning.

\paragraph{Execution Success vs.\ Workflow Graph Correctness.}
To directly analyze the alignment between execution outcomes and workflow graph construction, we perform a quadrant analysis for ES-ReAct, shown in Table~\ref{tab:graph_quality_quadrant}.

\begin{table}[ht!]
\centering
\small
\setlength{\tabcolsep}{5pt}
\begin{tabular}{lcc}
\toprule
\textbf{Configuration} & \textbf{Cases} & \textbf{Avg.\ Pass Rate} \\
\midrule
Execution $\checkmark$ + Graph $\checkmark$ & 44 & 89.5\% \\
Execution $\checkmark$ + Graph $\times$ & 71 & 0.0\% \\
Execution $\times$ + Graph $\checkmark$ & 3 & 46.7\% \\
Execution $\times$ + Graph $\times$ & 82 & 0.0\% \\
\bottomrule
\end{tabular}
\caption{Quadrant analysis of execution success and workflow graph correctness for ES-ReAct.}
\label{tab:graph_quality_quadrant}
\end{table}

The analysis reveals a clear asymmetry: execution success is a necessary but insufficient condition for end-to-end task success. Although many cases complete execution successfully, fewer than half of them are translated into correct workflow graphs, with failures primarily occurring during the summarization stage. Conversely, a small number of cases yield structurally correct workflows despite execution failures, achieving partial success. These findings are consistent with the mixed tool interference effects discussed in Section~\ref{sec:cognitive_burden_tool_interference}.

\paragraph{Discussion.}
This ablation study yields three key insights. First, the Execute-Summarize framework consistently improves small-model performance by decoupling execution from workflow graph construction. Second, strategy effectiveness is strongly task-dependent, with ES excelling in data-centric workflows and planning-based approaches better suited for logical reasoning. Third, workflow summarization emerges as the dominant bottleneck: even when execution succeeds, incorrect graph construction frequently prevents end-to-end success. Improving summarization fidelity is therefore a critical direction for future work.

\begin{table*}[ht!]
\centering
\small
\setlength{\tabcolsep}{6pt}
\begin{tabular}{llcccc}
\toprule
\textbf{Model} & \textbf{ES Variant} 
& \textbf{CRate} 
& \textbf{Exec. Compl.} 
& \textbf{Graph Valid.} 
& \textbf{Both Succ.} \\
\midrule
\multirow{4}{*}{Qwen3-8B}
& ReAct$\rightarrow$ReAct & 18.4 & 64.54 & 73.05 & 51.77 \\
& P\&E$\rightarrow$ReAct & 22.7 & 81.56 & 75.89 & 65.96 \\
& ReAct$\rightarrow$P\&E & 22.0 & 66.67 & 85.11 & 56.74 \\
& P\&E$\rightarrow$P\&E  & 21.3 & \textbf{88.65} & \textbf{87.23} & \textbf{78.72} \\
\midrule
\multirow{4}{*}{Qwen3-14B}
& ReAct$\rightarrow$ReAct & 21.3 & 74.47 & 92.20 & 71.63 \\
& P\&E$\rightarrow$ReAct & 19.1 & \textbf{90.07} & 97.16 & \textbf{88.65} \\
& ReAct$\rightarrow$P\&E & 23.4 & 75.89 & 90.07 & 69.50 \\
& P\&E$\rightarrow$P\&E  & \textbf{24.1} & 89.36 & \textbf{94.33} & 85.11 \\
\midrule
\multirow{4}{*}{Qwen3-32B}
& ReAct$\rightarrow$ReAct & 21.3 & 73.05 & 73.76 & 57.45 \\
& P\&E$\rightarrow$ReAct & 21.3 & \textbf{94.33} & 82.98 & 78.01 \\
& ReAct$\rightarrow$P\&E & 19.1 & 71.63 & 83.69 & 66.67 \\
& P\&E$\rightarrow$P\&E  & 19.1 & 89.36 & \textbf{87.23} & \textbf{79.43} \\
\midrule
\multirow{4}{*}{GPT-5}
& ReAct$\rightarrow$ReAct & 23.4 & 78.72 & \textbf{94.33} & 73.76 \\
& P\&E$\rightarrow$ReAct & 22.7 & 99.28 & 84.89 & \textbf{84.17} \\
& ReAct$\rightarrow$P\&E & \textbf{41.8} & 75.89 & 82.27 & 66.67 \\
& P\&E$\rightarrow$P\&E  & 35.5 & \textbf{100.0} & 80.85 & 80.85 \\
\bottomrule
\end{tabular}
\caption{
Detailed comparison of Execute--Summarize variants.
Only ES configurations are shown; single-stage baselines are reported in Table~\ref{tab:main_results}.
}
\label{tab:es_variants}
\end{table*}

\paragraph{Implications.}
The analyses in Tables~\ref{tab:141_overall}–\ref{tab:graph_quality_quadrant} reveal that workflow graph construction, rather than execution itself, is the dominant bottleneck for end-to-end success. While ES substantially improves execution reliability, failures frequently arise during summarization into valid and faithful graphs. This observation motivates future work on improving summarization fidelity and graph validation, beyond optimizing execution strategies alone.

\subsection{Detailed Results of Execute-Summarize Variants}
\label{app:es_detailed_results}

To complement the main results, we report detailed quantitative results for different Execute-Summarize (ES) strategy combinations. Unlike Table~\ref{tab:main_results}, this subsection focuses exclusively on ES variants
and omits single-stage baselines (ReAct and P\&E), allowing a clearer comparison of how execution and summarization strategies interact.

\paragraph{Execution-first vs.\ summarization-first effects.}
Table~\ref{tab:es_variants} shows that the execution strategy dominates execution completeness.
Across all models, P\&E-based execution consistently achieves the highest completeness
(e.g., 88.65\% on Qwen3-8B and 100\% on GPT-5),
while ReAct-based execution remains substantially lower regardless of summarization choice.

\paragraph{Graph validity as a summarization property.}
In contrast, graph validity is most sensitive to the summarization strategy.
ReAct$\rightarrow$P\&E and P\&E$\rightarrow$P\&E variants consistently outperform
their ReAct-summarized counterparts,
with gains of 10--15 points on smaller and mid-sized models.
This confirms that structured summarization induces more globally consistent workflows.

\paragraph{Joint success favors aligned strategies.}
The Both Success metric reveals that alignment between execution and summarization is crucial.
Fully aligned P\&E$\rightarrow$P\&E achieves the highest or near-highest joint success
on all Qwen models (+20--27 points over ReAct$\rightarrow$ReAct).
On GPT-5, while mixed strategies excel on individual metrics (e.g., ReAct$\rightarrow$P\&E on CRate),
P\&E$\rightarrow$P\&E remains the most balanced configuration when execution robustness
and graph fidelity are both required.

\paragraph{Guidance for strategy selection.}
These results suggest that:
(i) strengthening execution yields larger gains than strengthening summarization alone for smaller models;
(ii) summarization choice becomes increasingly important as execution quality saturates;
and (iii) fully aligned strategies provide the most stable performance across metrics and model scales.

\subsection{Computational Efficiency and Token Usage}
\label{app:efficiency}

A natural concern regarding the Execute--Summarize (ES) framework is whether the additional summarization stage introduces prohibitive computational overhead.
To examine this issue, we analyze average output token consumption across reasoning paradigms, model scales, and model families.

\begin{table}[ht]
\centering
\small
\setlength{\tabcolsep}{4pt}
\begin{tabular}{llcccc}
\toprule
\textbf{Model} & \textbf{Mode} & \textbf{Exec.} & \textbf{Sum.} & \textbf{Total} & \textbf{$\Delta$} \\
\midrule
\multirow{4}{*}{Qwen3-8B}
& ReAct    & 9,885  & --    & 9,885  & -- \\
& P\&E     & 14,132 & --    & 14,132 & -- \\
& ES-ReAct & 7,752  & 548   & 8,300  & $-$16.0\% \\
& ES-P\&E  & 8,777  & 547   & 9,324  & $-$34.0\% \\
\midrule
\multirow{4}{*}{Qwen3-14B}
& ReAct    & 9,841  & --    & 9,841  & -- \\
& P\&E     & 13,386 & --    & 13,386 & -- \\
& ES-ReAct & 8,535  & 661   & 9,196  & $-$6.6\% \\
& ES-P\&E  & 7,122  & 612   & 7,734  & $-$42.2\% \\
\midrule
\multirow{4}{*}{Qwen3-32B}
& ReAct    & 12,317 & --    & 12,317 & -- \\
& P\&E     & 16,119 & --    & 16,119 & -- \\
& ES-ReAct & 14,385 & 718   & 15,103 & $+$22.6\% \\
& ES-P\&E  & 8,273  & 725   & 8,998  & $-$44.2\% \\
\midrule
\multirow{4}{*}{GPT-4.1}
& ReAct    & 4,868  & --    & 4,868  & -- \\
& P\&E     & 4,521  & --    & 4,521  & -- \\
& ES-ReAct & 1,543  & 964   & 2,507  & $-$48.5\% \\
& ES-P\&E  & 696    & 575   & 1,271  & $-$71.9\% \\
\midrule
\multirow{4}{*}{GPT-5}
& ReAct    & 3,823  & --    & 3,823  & -- \\
& P\&E     & 10,013 & --    & 10,013 & -- \\
& ES-ReAct & 1,167  & 815   & 1,982  & $-$48.2\% \\
& ES-P\&E  & 1,775  & 1,005 & 2,780  & $-$72.2\% \\
\bottomrule
\end{tabular}
\caption{Average output token consumption per instance. Execution (Exec.) and summarization (Sum.) tokens are reported separately for ES variants. $\Delta$ denotes the relative change in total tokens compared to the corresponding single-stage baseline (ReAct or P\&E).}
\label{tab:token_efficiency}
\end{table}

\paragraph{Overall Token Efficiency.}
Table~\ref{tab:token_efficiency} shows that ES variants achieve substantial reductions in total output tokens across most models and reasoning paradigms, despite introducing an explicit summarization stage.
The effect is especially pronounced for ES-P\&E, which consistently reduces total token consumption by 34.0--44.2\% for Qwen models and 71.9--72.2\% for GPT models.
ES-ReAct also yields clear efficiency gains for GPT-4.1 and GPT-5 (approximately 48\%), while exhibiting more modest improvements for smaller Qwen models and a net overhead for Qwen3-32B.
Overall, these results indicate that stage separation does not introduce prohibitive cost; instead, it often leads to net efficiency gains.

\paragraph{Execution-Stage Reduction as the Primary Driver.}
The observed efficiency improvements are primarily driven by large reductions in execution-stage token usage.
By decoupling task execution from graph construction, ES eliminates tool-selection ambiguity and repeated reasoning loops caused by mixed tool invocation (MTI) patterns (Section~\ref{sec:cognitive_burden_tool_interference}), resulting in more focused and concise execution traces.
For instance, under GPT-4.1, ES-P\&E reduces execution tokens by approximately 85\% compared to standard P\&E (696 vs.\ 4,521 tokens), while Qwen3-14B achieves a 47\% reduction (7,122 vs.\ 13,386 tokens).
These execution-stage savings dominate the overall token budget and outweigh the additional summarization cost in most settings.

\paragraph{Summarization Overhead and Cross-Model Behavior.}
The summarization stage introduces a relatively small and stable overhead of roughly 547--725 tokens for Qwen models and 575--1,005 tokens for GPT models.
This cost remains limited because graph construction follows a deterministic schema (start node, function nodes, edges, and end node) and requires minimal deliberative reasoning.
Notably, ES-ReAct exhibits greater variance across model scales, including a token increase for Qwen3-32B, suggesting that free-form ReAct reasoning benefits less consistently from post-hoc summarization.
In contrast, ES-P\&E demonstrates robust and consistent efficiency gains across all evaluated models, indicating that structured planning provides a more stable foundation for effective stage separation.

\subsection{Diagnosing Performance Gaps Across Reasoning Paradigms}
\label{app:performance_gap_diagnosis}

To diagnose the sources of the performance gap between workflow generation paradigms, we conduct a fine-grained analysis of failure structures, token efficiency, and representative cases.
Rather than attributing the gap to model capacity, we show that it arises from systematic mismatches between reasoning structure and task requirements.

A central observation is that ReAct and P\&E exhibit strongly \textit{complementary} failure patterns, reflecting opposite weaknesses in how multiple objectives are handled within a single generation process.

\paragraph{ReAct: Intent Retention Failure.}
As shown in Table~\ref{tab:react_failures}, 30\% of ReAct failures correspond to cases where the model successfully executes the business logic but never constructs the workflow graph.
In these trajectories, the primary objective (execution) is completed, while the secondary objective (graph construction) is effectively forgotten.
An additional 61\% of failures involve partial completion of both objectives, indicating that maintaining multiple long-horizon objectives within a single reasoning chain is inherently unstable.

\paragraph{P\&E: Premature Abstraction.}
P\&E exhibits the opposite behavior.
In 84\% of failures, the model skips execution entirely and directly attempts to construct a workflow graph.
This premature abstraction yields graphs that are often structurally plausible but semantically incorrect, as they are inferred speculatively rather than grounded in actual tool execution.

\begin{table}[ht]
\centering
\small
\setlength{\tabcolsep}{4pt}
\begin{tabular}{lcccc}
\toprule
\textbf{Failure Mode} & \multicolumn{2}{c}{\textbf{ReAct}} & \multicolumn{2}{c}{\textbf{P\&E}} \\
\cmidrule(lr){2-3} \cmidrule(lr){4-5}
 & Count & \% & Count & \% \\
\midrule
Exec only, no graph   & 57  & 30.0 & 0   & 0.0 \\
Graph only, no exec   & 0   & 0.0  & 130 & 83.9 \\
Both partial/failed   & 116 & 61.1 & 24  & 15.5 \\
Empty/error           & 17  & 8.9  & 1   & 0.6 \\
\midrule
\textbf{Total Failures} & 190 & 100 & 155 & 100 \\
\bottomrule
\end{tabular}
\caption{Distribution of failure modes. ReAct predominantly fails by omitting graph construction, while P\&E fails by skipping execution and directly producing speculative graphs.}
\label{tab:react_failures}
\end{table}

Taken together, these complementary failures expose a fundamental limitation of single-stage multi-objective reasoning.
ReAct interleaves execution and abstraction within a single trajectory, making it vulnerable to intent degradation as execution lengthens.
P\&E, in contrast, exposes both business and graph tools during planning, enabling shortcut behaviors that bypass execution grounding altogether.
The ES framework resolves both issues by construction: execution is confined to business tools, while graph construction is deferred to a dedicated summarization stage with no access to execution tools.

These structural differences have direct implications for computational efficiency.
Table~\ref{tab:token_efficiency_detail} reports token consumption for both successful and failed cases.
ES variants consistently require fewer tokens, particularly on successful instances.
Notably, ES-P\&E uses only 3,672 tokens per successful case—less than one quarter of P\&E's 15,407 tokens—reflecting the elimination of verbose speculative reasoning.

\begin{table}[ht]
\centering
\small
\begin{tabular}{lcccc}
\toprule
\textbf{Mode} & \textbf{Succ.} & \textbf{Tokens/Succ.} & \textbf{Fail.} & \textbf{Tokens/Fail.} \\
\midrule
ReAct     & 10  & 8,310  & 190 & 9,695 \\
P\&E      & 45  & 15,407 & 155 & 13,438 \\
ES-ReAct  & 42  & 4,521  & 144 & 10,150 \\
ES-P\&E   & 50  & 3,672  & 129 & 10,500 \\
\bottomrule
\end{tabular}
\caption{Token consumption by outcome. ES variants substantially reduce token usage on successful cases and mitigate excessive waste on failures.}
\label{tab:token_efficiency_detail}
\end{table}

Aggregating across all failed instances, P\&E expends 2.08M tokens compared to 1.35M for ES-P\&E—a 35\% reduction.
This efficiency gain primarily stems from preventing long graph-only trajectories that dominate P\&E failures.

We further illustrate these differences with a representative case.
The task \texttt{math\_035\_iterative\_sqrt\_convergence} requires iterative numerical computation followed by correct workflow construction.
Table~\ref{tab:case_study} contrasts the behaviors of P\&E and ES-P\&E.

\begin{table}[ht]
\centering
\small
\begin{tabular}{lcc}
\toprule
\textbf{Metric} & \textbf{P\&E} & \textbf{ES-P\&E} \\
\midrule
Business actions & 0 & 3 (\texttt{sqrt}, \texttt{round\_to}, \texttt{finish}) \\
Graph actions & 15 & 8 \\
Total tokens & 16,244 & 3,170 \\
Pass rate & 0\% & 100\% \\
\bottomrule
\end{tabular}
\caption{Case study on \texttt{math\_035}. P\&E constructs a graph without execution, while ES-P\&E executes first and summarizes faithfully.}
\label{tab:case_study}
\end{table}

Under P\&E, the model immediately emits graph construction actions without invoking the required mathematical functions.
Although the resulting graph is structurally valid, it encodes incorrect semantics due to the absence of execution grounding.
In contrast, ES-P\&E enforces execution before abstraction, yielding both correctness and an 80\% reduction in token usage.

Overall, the performance gap between paradigms stems from structural misalignment rather than insufficient model capability.
ReAct and P\&E fail for opposite reasons—intent degradation versus premature abstraction—while ES resolves both through explicit stage separation.
These results highlight the importance of paradigm design in achieving both correctness and efficiency for complex workflow generation tasks.

\subsection{Execution-Summarization Interaction Analysis}
\label{app:execution_summarization_relationship}

We study how execution strategies and trace summarization choices affect the quality of generated workflow graphs under the Execute--Summarize (ES) paradigm.
ES decouples reasoning into two stages: (i) an \emph{execution stage}, where the model solves a task and produces execution traces, and (ii) a \emph{summarization stage}, where execution traces are distilled into a reusable workflow graph.
In our setting, the execution stage produces three independent rollouts, while the summarization stage constructs a graph based on selected execution traces.

We consider two execution strategies: \textbf{ReAct} and \textbf{Plan-and-Execute (P\&E)}, resulting in two ES variants, \textbf{ES-ReAct} and \textbf{ES-P\&E}.
To examine whether the choice of summarization sources matters, we evaluate three trace selection strategies: using all execution traces (\texttt{all}), using only successful traces (\texttt{all\_success}), and randomly sampling one successful trace (\texttt{random\_success}).
All experiments are conducted on the subset of 141 tasks that are solvable by a strong Claude baseline, ensuring that failures primarily reflect execution--summarization interactions rather than task infeasibility.

\paragraph{Overall performance.}
Table~\ref{tab:es_overall_pass_rate} reports both case-level and test-level pass rates under different ES configurations.
Across all trace selection strategies, ES-P\&E consistently outperforms ES-ReAct.
The best configuration, ES-P\&E with \texttt{random\_success} selection, achieves a case pass rate of 24.82\% and a test pass rate of 30.26\%.
However, the performance gap among different trace selection strategies remains small (within 2\%), suggesting that ES is relatively insensitive to the specific choice of summarization sources.

\begin{table*}[t]
\centering
\small
\setlength{\tabcolsep}{4pt}
\begin{tabular}{lcc|cc}
\toprule
\textbf{Setting} 
& \textbf{Case Pass Rate} & \textbf{Solved}
& \textbf{Test Pass Rate} & \textbf{Passed Tests} \\
\midrule
ES-ReAct (all)              
& 24.11\% & 34/141 
& 28.67\% & 199/694 \\

ES-ReAct (all\_success)    
& 23.40\% & 33/141 
& 28.24\% & 196/694 \\

ES-ReAct (random\_success) 
& 23.40\% & 33/141 
& 28.67\% & 199/694 \\
\midrule
ES-P\&E (all)              
& \textbf{25.53\%} & 36/141 
& 30.12\% & 209/694 \\

ES-P\&E (all\_success)     
& 24.11\% & 34/141 
& 27.38\% & 190/694 \\

ES-P\&E (random\_success)  
& 24.82\% & 35/141 
& \textbf{30.26\%} & 210/694 \\
\bottomrule
\end{tabular}
\caption{Case-level and test-level performance under the Execute--Summarize (ES) paradigm with different trace selection strategies.}
\label{tab:es_overall_pass_rate}
\end{table*}

\paragraph{Single Execute--Summarize vs.\ multi-trace ES.}
To isolate the contribution of trace aggregation, we further compare the above results with a simplified setting where the model performs only a single Execute--Summarize pass, i.e., one execution followed by one summarization, without any trace selection.
As shown in Table~\ref{tab:es_single_pass_test}, even a single ES pass substantially improves over non-ES baselines, achieving test pass rates of 23.2\% (ES-ReAct) and 25.9\% (ES-P\&E).

\begin{table}[t]
\centering
\small
\setlength{\tabcolsep}{6pt}
\begin{tabular}{lcc}
\toprule
\textbf{Model} & \textbf{Mode} & \textbf{Test Pass Rate} \\
\midrule
Qwen3-8B & ES-ReAct (single ES pass) & 23.2\% \\
Qwen3-8B & ES-P\&E  (single ES pass) & \textbf{25.9}\% \\
\bottomrule
\end{tabular}
\caption{Test pass rates of a single Execute--Summarize pass without trace selection.}
\label{tab:es_single_pass_test}
\end{table}

Comparing Tables~\ref{tab:es_overall_pass_rate} and~\ref{tab:es_single_pass_test}, we observe that a single ES cycle already captures most of the gains brought by the ES paradigm.
While aggregating multiple execution traces provides additional improvements, the marginal benefit is limited.
This indicates that the primary advantage of ES arises from explicit stage separation and summarization, rather than from complex trace selection heuristics.

\paragraph{Insights and implications.}
Overall, these results suggest that (i) Plan-and-Execute is a more effective execution strategy within ES, (ii) ES is robust to the choice of summarization sources, and (iii) even a minimal ES configuration is sufficient to realize most of the performance gains.
From a practical perspective, this makes ES attractive as a lightweight yet effective framework: practitioners can adopt a simple single-pass ES or a low-cost \texttt{random\_success} strategy without sacrificing much performance.

\paragraph{Summary.}
Taken together, the results in Tables~\ref{tab:es_overall_pass_rate} and~\ref{tab:es_single_pass_test} demonstrate that the core benefit of ES stems from explicit stage separation rather than sophisticated trace aggregation. Plan-and-Execute consistently serves as a stronger execution backbone, while ES remains robust to the choice of summarization sources.
These findings motivate our default use of ES-P\&E with lightweight trace selection in the main experiments (Section~\ref{sec:main_results}).

\section{Details of Experiment Settings}
\label{app:experiment_settings}
\subsection{Model Size and Budget}
\label{app:model_size_and_budget}

We evaluate five models spanning both proprietary APIs and publicly documented open-weight models: \texttt{gpt-4.1}, \texttt{gpt-5}, \texttt{Qwen3-8B}, \texttt{Qwen3-14B}, and \texttt{Qwen3-32B}. For proprietary models (i.e., \texttt{gpt-4.1} and \texttt{gpt-5}), we access them via the Azure OpenAI (AOAI) service. As is common for commercial API-based models, vendors do not disclose exact parameter counts or full training specifications. Consequently, we report these models only by their official names and version identifiers, without making claims about their parameter sizes.

For models with publicly available specifications, namely the Qwen3 series, we report model sizes according to the official releases: 8B, 14B, and 32B parameters, respectively. These models are deployed locally on GPU servers equipped with NVIDIA A100 GPUs (with additional experiments conducted on H100 GPUs). This setup enables a controlled comparison between proprietary and open models under realistic deployment conditions, while respecting the differing levels of transparency in model specifications.

\paragraph{Execution Budget and Runtime Accounting.}
\begin{table}[ht]
\centering
\small
\setlength{\tabcolsep}{6pt}
\begin{tabular}{lcc}
\toprule
\textbf{Model} & \textbf{Total Time} & \textbf{Days} \\
\midrule
Qwen3-8B  & 221h 8m  & 9.2  \\
Qwen3-14B & 464h 32m & 19.4 \\
Qwen3-32B & 490h 41m & 20.4 \\
GPT-4.1   & 21h 54m  & 0.9  \\
GPT-5     & 18h 54m  & 0.8  \\
\bottomrule
\end{tabular}
\caption{Total cumulative execution time across all cases and all six experimental configurations for each model.}
\label{tab:total_time}
\end{table}
We further report the execution budget of our experiments in terms of cumulative model invocation time. For each model, we run six experimental configurations: standard \textit{ReAct}, standard \textit{Plan-and-Execute (P\&E)}, and four FlowMind variants that combine the two strategies across stages, namely \textit{ReAct--ReAct (RR)}, \textit{Plan--ReAct (PR)}, \textit{ReAct--Plan (RP)}, and \textit{Plan--Plan (PP)}.

Table~\ref{tab:total_time} summarizes the total execution time aggregated across all cases and all six configurations for each model, while Table~\ref{tab:mode_time} further breaks down the runtime by execution mode.

\begin{table*}[ht!]
\centering
\small
\setlength{\tabcolsep}{6pt}
\begin{tabular}{lccccc}
\toprule
\textbf{Mode} & \textbf{8B} & \textbf{14B} & \textbf{32B} & \textbf{GPT-4.1} & \textbf{GPT-5} \\
\midrule
ReAct (R) & 46.5 & 110.4 & 92.9 & 8.1 & 4.2 \\
P\&E (P)  & 56.1 & 129.6 & 133.8 & 7.6 & 10.8 \\
ES-RR        & 24.6 & 58.9  & 48.7 & 1.4 & 0.8 \\
ES-PR        & 31.0 & 77.5  & 81.3 & 1.3 & 1.1 \\
ES-RP        & 26.7 & 36.7  & 51.8 & 1.4 & 0.8 \\
ES-PP        & 36.3 & 51.5  & 82.1 & 2.1 & 1.2 \\
\midrule
Total     & 221.1 & 464.5 & 490.7 & 21.9 & 18.9 \\
\bottomrule
\end{tabular}
\caption{Execution time (in hours) by model and strategy. For each model, six strategies are evaluated: ReAct (R), Plan \& Execute (P), and four Execute--Summarize (ES) variants. ES-XY denotes the combination of strategies (R or P) used in the Execution (X) and Summarization (Y) stages. Reported values are cumulative execution times over all evaluation tasks.}
\label{tab:mode_time}
\end{table*}

The reported execution time corresponds to the serially accumulated runtime across all cases, rather than wall-clock time. For GPT models, runtime is measured based on AOAI API invocation timestamps, whereas for Qwen3 models, time is recorded from local system execution logs. These measurements include system-level overhead such as retry logic, scheduling latency, and orchestration costs; as a result, the effective per-call inference time is shorter than the reported totals.

We emphasize that the execution budget is inherently dependent on the underlying hardware configuration and service provider characteristics. The statistics reported above reflect a specific deployment setting, with Qwen3 models running on A100-class GPUs and GPT models served via AOAI. In practice, runtime is sensitive to multiple external factors, including GPU type, service-side rate limits, request scheduling policies, batching strategies, and the degree of concurrency during inference. Different infrastructure setups or service providers may therefore yield substantially different runtime profiles under otherwise identical workloads.

Finally, we note that the summarized execution budget accounts only for the finalized experimental runs reported in this paper. Due to extensive research iterations---including prompt refinement, hyperparameter tuning, and multiple ablation studies---the overall computational and API budget incurred during development is substantially higher than the figures reported above.

\subsection{Packages and Parameter Settings}
\label{app:env_and_params}

Our preprocessing, prompting, and evaluation pipeline is primarily implemented using custom scripts, rather than third-party NLP toolkits (e.g., NLTK, SpaCy, or ROUGE), for tokenization, normalization, or metric computation. This design choice minimizes hidden preprocessing assumptions and ensures full control over execution behavior. For answer extraction, models are instructed to produce outputs in a fixed and explicitly specified format, and we employ regular-expression matching to parse the final predicted label from the generated text.

\paragraph{Model Deployment Environment.}
Local model inference is conducted using the \texttt{vLLM} serving framework. Table~\ref{tab:env_vllm} summarizes the software stack used in our experiments. Most experiments are executed on servers equipped with $4 \times$ NVIDIA A100 (80GB PCIe) GPUs, while a smaller subset is conducted on machines with $4 \times$ NVIDIA H100 GPUs. No experiment relies on hardware-specific optimizations beyond standard CUDA and framework defaults.

\begin{table}[ht]
\centering
\small
\setlength{\tabcolsep}{6pt}
\begin{tabular}{l l}
\toprule
\textbf{Component} & \textbf{Version} \\
\midrule
Python       & 3.10.19 \\
PyTorch      & 2.8.0+cu128 \\
CUDA         & 12.8 \\
Transformers & 4.57.3 \\
vLLM         & 0.11.0 \\
\bottomrule
\end{tabular}
\caption{Software environment for vLLM-based model deployment.}
\label{tab:env_vllm}
\end{table}

\paragraph{API-based Models.}
For proprietary models accessed via hosted services, we use the official OpenAI API, including \texttt{gpt-4.1} and \texttt{gpt-5}. Since the underlying hardware configuration, serving stack, and request-level concurrency control are managed by the service provider, these factors are not explicitly controlled in our experiments.

\paragraph{Decoding and Reproducibility.}
Our preprocessing, prompting, and evaluation pipeline is primarily implemented with custom scripts rather than third-party NLP toolkits (e.g., NLTK, SpaCy, or ROUGE) for tokenization, normalization, and metric computation. For answer extraction, models are instructed to produce outputs in a fixed format, and regular-expression matching is used to parse the final predicted label from generated text.
Decoding parameters (e.g., temperature, maximum tokens, and stopping criteria) are fixed per experiment and recorded in configuration files.  To support reproducibility, the released code and configuration files record the software environment and dependency versions (including Python and any SDKs or client libraries used for model invocation), together with all experiment-specific decoding and runtime settings. Additional low-level implementation details are provided in the accompanying source code.

\section{Prompts}
The tables below provide a comprehensive list of the prompts used in this work.

\onecolumn
\clearpage


\begin{table*}[t]
\centering
\scriptsize
\renewcommand\arraystretch{1.0}

\begin{tabular}{p{0.95\linewidth}}
\toprule[1.5pt]
\textbf{System Prompt: ReAct Strategy} \\
\midrule[1pt]

\textbf{Strategy Overview:} The ReAct (Reasoning and Acting) pattern interleaves reasoning traces with action execution. In each iteration, the LLM: (1) reasons about the current state, (2) selects and executes a tool, (3) observes the result, and (4) repeats until task completion via the \texttt{finish} action. \\
\midrule

\textbf{Prompt Content:} \\
\texttt{You are a helpful AI assistant using the ReAct (Reasoning and Acting) framework.} \\
\\
\texttt{You have access to the following functions:} \\
\texttt{\{functions\_text\}} \\
\\
\texttt{\#\# Your Task} \\
\texttt{You will be given a problem that may include example inputs and expected outputs.} \\
\texttt{You MUST use the available functions to compute results from the given input data.} \\
\\
\texttt{\#\# CRITICAL REQUIREMENT - Process Examples with Functions} \\
\texttt{- You MUST call business functions (not just 'finish') to process the given input data} \\
\texttt{- Work through at least one example by calling the appropriate functions} \\
\texttt{- Continue calling functions until you compute the expected output} \\
\texttt{- Only call 'finish' AFTER you have used functions to process the input and obtained results} \\
\\
\texttt{Example workflow:} \\
\texttt{Given: "Calculate sum of [1,2,3]" with expected output 6} \\
\texttt{1. Call add/sum function with appropriate input -> Get result: 6} \\
\texttt{2. Verify result matches expected output} \\
\texttt{3. Call 'finish' with the computed answer "6"} \\
\\
\texttt{**FORBIDDEN:** Calling 'finish' immediately without first using business functions to process input} \\
\texttt{**REQUIRED:** At least one business function call before 'finish'} \\
\\
\texttt{CRITICAL: You MUST respond ONLY with valid JSON in the exact format shown below.} \\
\texttt{Do NOT add any explanation outside the JSON.} \\
\\
\texttt{Required JSON format for EVERY response:} \\
\texttt{\{} \\
\hspace*{1em}\texttt{"reasoning": "Your thought process about what to do next",} \\
\hspace*{1em}\texttt{"action": "function\_name",} \\
\hspace*{1em}\texttt{"action\_input": \{} \\
\hspace*{2em}\texttt{"parameter\_name": value} \\
\hspace*{1em}\texttt{\}} \\
\texttt{\}} \\
\\
\texttt{IMPORTANT RULES:} \\
\texttt{1. ALWAYS respond with valid JSON - no other text before or after} \\
\texttt{2. Use available functions to solve tasks - don't try to solve manually} \\
\texttt{3. Refer to the function definitions above for correct parameter names} \\
\texttt{4. When done, call 'finish' function with your answer} \\
\texttt{5. Each response must be a single JSON object} \\
\\
\texttt{When you have the final answer:} \\
\texttt{\{} \\
\hspace*{1em}\texttt{"reasoning": "I have processed the input with functions and obtained the result",} \\
\hspace*{1em}\texttt{"action": "finish",} \\
\hspace*{1em}\texttt{"action\_input": \{} \\
\hspace*{2em}\texttt{"answer": "Your computed answer here"} \\
\hspace*{1em}\texttt{\}} \\
\texttt{\}} \\
\\
\texttt{REMEMBER:} \\
\texttt{- Output ONLY valid JSON, nothing else} \\
\texttt{- Use the functions provided - they are the correct way to solve tasks} \\
\texttt{- Process at least one example input before finishing} \\
\texttt{- The 'finish' function is the ONLY way to output your final answer} \\
\bottomrule[1.5pt]
\end{tabular}%
\caption{ReAct strategy system prompt. The LLM must output structured JSON containing \texttt{reasoning}, \texttt{action}, and \texttt{action\_input} fields. The workflow enforces tool usage before the terminal \texttt{finish} action.}
\label{tab:react_system_prompt}
\end{table*}


\begin{table*}[t]
\centering
\scriptsize
\renewcommand\arraystretch{1.0}

\begin{tabular}{p{0.95\linewidth}}
\toprule[1.5pt]
\textbf{System Prompt: Plan-and-Execute Strategy (Planning Phase)} \\
\midrule[1pt]

\textbf{Strategy Overview:} Unlike ReAct's interleaved approach, Plan-and-Execute separates planning from execution. In the planning phase, the LLM analyzes the task and generates a complete step-by-step plan \textit{without} executing any tools. This plan is then executed sequentially in the execution phase. \\
\midrule

\textbf{Prompt Content:} \\
\texttt{You are a strategic planner. Analyze the task and create a step-by-step execution plan.} \\
\\
\texttt{Available functions that will be used during execution:} \\
\texttt{\{functions\_text\}} \\
\\
\texttt{IMPORTANT: This is the PLANNING phase only. Do NOT call any functions.} \\
\texttt{Just analyze the task and output a plan.} \\
\\
\texttt{\#\# CRITICAL REQUIREMENT - Process Examples with Functions} \\
\texttt{Your plan MUST include steps that use business functions to compute results from example inputs.} \\
\texttt{- Each step should specify which function to call} \\
\texttt{- The plan must process at least one example input to produce the expected output} \\
\texttt{- Do NOT plan to skip computation - you MUST use functions to process input data} \\
\\
\texttt{**FORBIDDEN:** Planning to directly return an answer without using functions to compute it} \\
\texttt{**REQUIRED:** At least one step that uses a business function to process input data} \\
\\
\texttt{PLANNING GUIDELINES:} \\
\texttt{1. Break down the task into clear, atomic steps} \\
\texttt{2. Each step should accomplish ONE specific sub-goal using a function} \\
\texttt{3. Consider the order of operations and dependencies} \\
\texttt{4. Keep the plan concise (typically 2-5 steps)} \\
\texttt{5. The final step should return the computed answer} \\
\\
\texttt{You MUST respond with valid JSON in this exact format:} \\
\texttt{\{} \\
\hspace*{1em}\texttt{"analysis": "Brief analysis of what needs to be done",} \\
\hspace*{1em}\texttt{"steps": [} \\
\hspace*{2em}\texttt{"Step 1: Use function\_name to do something",} \\
\hspace*{2em}\texttt{"Step 2: Use another\_function to do something else",} \\
\hspace*{2em}\texttt{"Step 3: Return the computed result"} \\
\hspace*{1em}\texttt{]} \\
\texttt{\}} \\
\\
\texttt{Output ONLY valid JSON, nothing else.} \\
\bottomrule[1.5pt]
\end{tabular}%
\caption{Plan-and-Execute strategy planning phase prompt. The LLM outputs a structured plan as JSON with \texttt{analysis} and \texttt{steps} fields, without executing any tools during this phase.}
\label{tab:plan_execute_planning_prompt}
\end{table*}


\begin{table*}[t]
\centering
\scriptsize
\renewcommand\arraystretch{1.0}

\begin{tabular}{p{0.95\linewidth}}
\toprule[1.5pt]
\textbf{System Prompt: Plan-and-Execute Strategy (Execution Phase)} \\
\midrule[1pt]

\textbf{Phase Description:} After Planning phase generates a step-by-step plan, this phase executes each step using ReAct-style tool calling. The plan analysis is provided as context, and each step receives accumulated results from previous steps. \\
\midrule

\textbf{System Prompt Content:} \\
\texttt{You are a helpful AI assistant executing a pre-planned task.} \\
\\
\texttt{You have access to the following functions:} \\
\texttt{\{functions\_text\}} \\
\\
\texttt{\#\# CRITICAL REQUIREMENT - Process Input with Functions} \\
\texttt{You are executing a step to process input data toward the expected output.} \\
\texttt{- You MUST use the available business functions to compute results} \\
\texttt{- Do NOT skip to 'finish' without actually calling functions to process data} \\
\texttt{- Each step should use at least one business function before calling 'finish'} \\
\\
\texttt{You MUST respond with valid JSON in this exact format:} \\
\texttt{\{"reasoning": "Your thought process", "action": "function\_name",} \\
\texttt{ "action\_input": \{"parameter\_name": value\}\}} \\
\\
\texttt{RULES:} \\
\texttt{1. Focus on completing the CURRENT STEP described in the task} \\
\texttt{2. Use the available business functions to accomplish the step} \\
\texttt{3. Refer to the function definitions above for correct parameter names} \\
\texttt{4. Call 'finish' with the computed result when done} \\
\texttt{5. Output ONLY valid JSON, nothing else} \\
\midrule

\textbf{Plan Context (Injected as Initial User Message):} \\
The complete plan analysis from Planning phase is provided as the first user message, giving the LLM context about the overall task structure. \\
\midrule

\textbf{Step Instruction Template (per step):} \\
\texttt{Now execute Step \{step\_number\}: \{step\_description\}} \\
\\
\texttt{Previous results:} \\
\texttt{\{accumulated\_step\_context\}} \\
\\
\texttt{Execute ALL steps described in the analysis above. Only call 'finish' with the} \\
\texttt{FINAL computed result after completing all steps.} \\
\bottomrule[1.5pt]
\end{tabular}%
\caption{Plan-and-Execute strategy execution phase prompt. The plan from Planning phase is injected as initial context, and each step receives its description plus accumulated results from previous steps.}
\label{tab:plan_execute_execution_prompt}
\end{table*}


\begin{table*}[t]
\centering
\scriptsize
\renewcommand\arraystretch{1.0}

\begin{tabular}{p{0.95\linewidth}}
\toprule[1.5pt]
\textbf{Execute-Summarize Strategy Framework Overview} \\
\midrule[1pt]

\textbf{Strategy Description:} Execute-Summarize is a two-stage orchestration pattern: \\
\hspace*{1em}1. \textbf{Execute Stage}: Solve business problems using domain-specific tools \\
\hspace*{1em}2. \textbf{Summarize Stage}: Convert execution traces into reusable workflow graphs \\
\\
Each stage can independently use different orchestration strategies, tool sets, and LLM configurations. \\
\midrule

\textbf{Execute Stage - Strategy Options:} \\
\texttt{class OrchestratorStrategy(Enum):} \\
\hspace*{1em}\texttt{REACT = "react"} \hspace*{6em}-- Interleaved reasoning-action loop \\
\hspace*{1em}\texttt{PLAN\_AND\_EXECUTE = "plan\_and\_execute"} -- Plan first, then execute \\
\midrule

\textbf{Summarize Stage (Graph Building) - Strategy Options:} \\
\texttt{class GraphBuildStrategy(Enum):} \\
\hspace*{1em}\texttt{REACT = "react"} \hspace*{6em}-- Iterative graph construction \\
\hspace*{1em}\texttt{PLAN\_AND\_BUILD = "plan\_and\_build"} \hspace*{1em}-- Plan graph structure, then build \\
\midrule

\textbf{Strategy Combinations (2 $\times$ 2 = 4 configurations):} \\
\begin{tabular}{l|cc}
 & \textbf{Summarize: ReAct} & \textbf{Summarize: Plan-and-Build} \\
\hline
\textbf{Execute: ReAct} & ReAct + ReAct & ReAct + Plan-and-Build \\
\textbf{Execute: Plan-and-Execute} & Plan-and-Execute + ReAct & Plan-and-Execute + Plan-and-Build \\
\end{tabular} \\
\bottomrule[1.5pt]
\end{tabular}%
\caption{Execute-Summarize strategy framework overview. Both Execute and Summarize stages support strategy selection, enabling 4 possible configurations for workflow generation.}
\label{tab:execute_summarize_overview}
\end{table*}


\begin{table*}[t]
\centering
\scriptsize
\renewcommand\arraystretch{1.0}

\begin{tabular}{p{0.95\linewidth}}
\toprule[1.5pt]
\textbf{System Prompt: Summarize Stage - Graph Building Implementation} \\
\midrule[1pt]

\textbf{Purpose:} When the Summarize stage is configured to build workflow graphs, this system prompt guides the LLM to convert execution traces into reusable graph structures through iterative tool calling. \\
\midrule

\textbf{Prompt Content:} \\
\texttt{You are a Graph Building Agent that converts execution traces into workflow graphs.} \\
\\
\texttt{\#\# Your Task} \\
\texttt{You will be given execution trace(s) from completed tasks. Your job is to build a workflow} \\
\texttt{graph that represents this execution flow. The graph should:} \\
\texttt{1. Capture the data flow between function calls} \\
\texttt{2. Use appropriate node types (START, END, FUNCTION\_CALL, CONDITION, LLM)} \\
\texttt{3. Connect nodes with proper edges} \\
\\
\texttt{\#\# Response Format} \\
\texttt{ALWAYS respond with valid JSON only:} \\
\texttt{\{"reasoning": "Your analysis and what you plan to do",} \\
\texttt{ "action": "tool\_name", "action\_input": \{"param1": "value1"\}\}} \\
\\
\texttt{\#\# Graph Building Process} \\
\texttt{1. Start: Always begin with `add\_start\_node` to initialize the workflow} \\
\texttt{2. Function Nodes: Add function nodes for each function call in the trace} \\
\texttt{3. Connections: Add edges to connect all nodes in execution order} \\
\texttt{4. End: Add `add\_end\_node` and connect it} \\
\texttt{5. Complete: Call `finish` when finished} \\
\\
\texttt{\#\# Important Rules} \\
\texttt{1. JSON Only: ALWAYS respond with valid JSON only. No other text.} \\
\texttt{2. One Action Per Response: Call exactly ONE tool per response} \\
\texttt{3. Data Flow: Ensure output\_key of one node matches input\_keys of downstream nodes} \\
\texttt{4. Connect Everything: Every node must have edges} \\
\\
\texttt{\{tools\_prompt\}} \\
\\
\texttt{\#\# Pattern Recognition} \\
\texttt{- Linear Pattern: Functions called in sequence without repetition} \\
\texttt{- Loop Pattern: Same function called multiple times with termination condition} \\
\texttt{  - Create condition\_node with condition\_expr for termination check} \\
\texttt{  - Add loop-back edge from loop body to condition\_node} \\
\texttt{- Parallel Pattern: Multiple independent function calls} \\
\bottomrule[1.5pt]
\end{tabular}%
\caption{Summarize stage system prompt for Graph Building implementation. The LLM iteratively calls graph manipulation tools to construct workflow graphs from execution traces.}
\label{tab:summarize_graph_system}
\end{table*}


\begin{table*}[t]
\centering
\scriptsize
\renewcommand\arraystretch{1.0}

\begin{tabular}{p{0.95\linewidth}}
\toprule[1.5pt]
\textbf{System Prompt: Summarize Stage - Graph Building with Plan-and-Execute (Planning Phase)} \\
\midrule[1pt]

\textbf{Purpose:} When Graph Building uses Plan-and-Execute mode, this prompt guides the LLM to first create a construction plan before building. The LLM analyzes execution traces and outputs a step-by-step plan without calling any tools. \\
\midrule

\textbf{Prompt Content:} \\
\texttt{You are a Graph Building Planner. Your job is to analyze execution traces and create a} \\
\texttt{step-by-step plan for building a workflow graph.} \\
\\
\texttt{\#\# Your Task} \\
\texttt{You will be given execution trace(s) from completed tasks. Analyze them and create a} \\
\texttt{detailed plan for building the workflow graph.} \\
\\
\texttt{IMPORTANT: This is the PLANNING phase only. Do NOT call any tools.} \\
\texttt{Just analyze and output a plan.} \\
\\
\texttt{\#\# Available Graph Building Tools (for reference only - do not call them now)} \\
\texttt{\{tools\_prompt\}} \\
\\
\texttt{\#\# Planning Guidelines} \\
\texttt{1. Analyze the trace: Understand the data flow and function call sequence} \\
\texttt{2. Identify patterns: Linear, loop, or parallel patterns} \\
\texttt{3. Plan nodes: List all nodes needed (start, function nodes, condition nodes, end)} \\
\texttt{4. Plan edges: List all connections between nodes} \\
\texttt{5. Consider data flow: Ensure output\_key of one node matches input\_keys of downstream nodes} \\
\\
\texttt{\#\# Output Format} \\
\texttt{You MUST respond with valid JSON in this exact format:} \\
\texttt{\{} \\
\hspace*{1em}\texttt{"analysis": "Brief analysis of what the trace shows and what pattern it represents",} \\
\hspace*{1em}\texttt{"steps": [} \\
\hspace*{2em}\texttt{"Step 1: Add start node with initial data containing ...",} \\
\hspace*{2em}\texttt{"Step 2: Add function node for ... with input\_keys mapping ...",} \\
\hspace*{2em}\texttt{"Step 3: Add edge from ... to ...",} \\
\hspace*{2em}\texttt{"...",} \\
\hspace*{2em}\texttt{"Step N: Call finish to complete graph building"} \\
\hspace*{1em}\texttt{]} \\
\texttt{\}} \\
\\
\texttt{Output ONLY valid JSON, nothing else.} \\
\bottomrule[1.5pt]
\end{tabular}%
\caption{Summarize stage planning prompt for Graph Building with Plan-and-Execute mode. The LLM creates a structured plan for graph construction without executing any tools.}
\label{tab:summarize_graph_planning}
\end{table*}


\begin{table*}[t]
\centering
\scriptsize
\renewcommand\arraystretch{1.0}

\begin{tabular}{p{0.95\linewidth}}
\toprule[1.5pt]
\textbf{System Prompt: Summarize Stage - Graph Building with Plan-and-Execute (Execution Phase)} \\
\midrule[1pt]

\textbf{Purpose:} After the planning phase produces a step-by-step plan, this prompt guides the LLM to execute each step by calling the appropriate graph manipulation tools. The focus is on faithful execution of the pre-generated plan. \\
\midrule

\textbf{Prompt Content:} \\
\texttt{You are a Graph Building Agent executing a pre-planned graph construction.} \\
\\
\texttt{\#\# Your Task} \\
\texttt{Execute the steps in the plan to build the workflow graph. Call the appropriate tools} \\
\texttt{for each step.} \\
\\
\texttt{\#\# Available Graph Building Tools} \\
\texttt{\{tools\_prompt\}} \\
\\
\texttt{\#\# Response Format} \\
\texttt{ALWAYS respond with valid JSON only:} \\
\texttt{\{} \\
\hspace*{1em}\texttt{"reasoning": "What step you are executing and why",} \\
\hspace*{1em}\texttt{"action": "tool\_name",} \\
\hspace*{1em}\texttt{"action\_input": \{} \\
\hspace*{2em}\texttt{"param1": "value1"} \\
\hspace*{1em}\texttt{\}} \\
\texttt{\}} \\
\\
\texttt{\#\# Rules} \\
\texttt{1. Follow the plan: Execute steps in order as planned} \\
\texttt{2. JSON Only: ALWAYS respond with valid JSON only} \\
\texttt{3. One Action Per Response: Call exactly ONE tool per response} \\
\texttt{4. Complete All Steps: Execute all steps before calling finish} \\
\\
\texttt{When you have completed ALL steps in the plan:} \\
\texttt{\{} \\
\hspace*{1em}\texttt{"reasoning": "All steps in the plan have been executed, graph is complete",} \\
\hspace*{1em}\texttt{"action": "finish",} \\
\hspace*{1em}\texttt{"action\_input": \{} \\
\hspace*{2em}\texttt{"answer": "Description of the completed graph"} \\
\hspace*{1em}\texttt{\}} \\
\texttt{\}} \\
\bottomrule[1.5pt]
\end{tabular}%
\caption{Summarize stage execution prompt for Graph Building with Plan-and-Execute mode. The LLM executes the pre-planned graph construction steps by calling graph manipulation tools.}
\label{tab:summarize_graph_execution}
\end{table*}


\begin{table*}[t]
\centering
\scriptsize
\renewcommand\arraystretch{1.0}

\begin{tabular}{p{0.95\linewidth}}
\toprule[1.5pt]
\textbf{User Message Template: Execution Trace Input for Graph Building (Summarize Stage)} \\
\midrule[1pt]

\textbf{Purpose:} This template formats execution traces from the Execute stage as input for Graph Building. It provides the LLM with structured information about each function call, its arguments, and outputs. \\
\midrule

\textbf{Template Content:} \\
\texttt{\#\# Execution Trace(s) to Convert} \\
\\
\texttt{\#\#\# Execution \{i\}: \{task\_id\}} \\
\texttt{**Query**: \{query\}} \\
\texttt{**Status**: Success / Failed} \\
\texttt{**Final Answer**: \{final\_answer\}} \\
\\
\texttt{**Function Calls**:} \\
\texttt{- Step 1: `\{action\}(\{args\_str\})` -> `\{observation\}`} \\
\texttt{- Step 2: `\{action\}(\{args\_str\})` -> `\{observation\}`} \\
\texttt{...} \\
\\
\texttt{---} \\
\texttt{Now build a workflow graph that represents this execution.} \\
\texttt{Start by adding the start node, then add function nodes for each step.} \\
\texttt{Respond with JSON only.} \\
\bottomrule[1.5pt]
\end{tabular}%
\caption{Execution trace input template for Graph Building in Summarize stage. Provides structured trace information from the Execute stage to guide graph construction.}
\label{tab:summarize_graph_input}
\end{table*}


\begin{table*}[t]
\centering
\scriptsize
\renewcommand\arraystretch{1.0}

\begin{tabular}{p{0.95\linewidth}}
\toprule[1.5pt]
\textbf{User Message Template: Graph Building Observation Feedback (Summarize Stage)} \\
\midrule[1pt]

\textbf{Purpose:} After each graph manipulation tool execution, this observation message provides feedback to the LLM. It includes the tool result (or error), current graph state, and missing elements (nodes without proper edges). \\
\midrule

\textbf{Template Content:} \\
\texttt{\#\# Observation} \\
\texttt{\{result\_str\}} \\
\\
\texttt{\{graph\_state\}} \\
\\
\texttt{\{missing\_info\}} \\
\\
\texttt{What's your next action? (respond with JSON)} \\
\midrule

\textbf{Graph State Example:} \\
\texttt{Nodes (3):} \\
\hspace*{1em}\texttt{- \_\_start\_\_ (START)} \\
\hspace*{1em}\texttt{- step1\_add (BUSINESS): add(a=num1, b=num2) -> sum\_result} \\
\hspace*{1em}\texttt{- \_\_end\_\_ (END)} \\
\\
\texttt{Edges (2):} \\
\hspace*{1em}\texttt{- \_\_start\_\_ -> step1\_add} \\
\hspace*{1em}\texttt{- step1\_add -> \_\_end\_\_} \\
\\
\texttt{Missing Elements: None} \\
\bottomrule[1.5pt]
\end{tabular}%
\caption{Graph building observation feedback template in Summarize stage. Provides real-time graph state visualization and identifies missing connections after each tool execution.}
\label{tab:summarize_graph_observation}
\end{table*}


\begin{table*}[t]
\centering
\scriptsize
\renewcommand\arraystretch{1.0}

\begin{tabular}{p{0.95\linewidth}}
\toprule[1.5pt]
\textbf{System Prompt: Function Call Error Recovery} \\
\midrule[1pt]

\textbf{Purpose:} When a tool execution fails (e.g., parameter name mismatch, type errors, or schema violations), this prompt provides error context to enable LLM self-correction. It includes the failed call details, error message, correct function signature, and retry count. \\
\midrule

\textbf{Prompt Content:} \\
\texttt{\#\# Execution Error} \\
\\
\texttt{The previous step failed. Please analyze the error and try again.} \\
\\
\texttt{**Function called**: \{function\_name\}} \\
\texttt{**Arguments you used**: \{args\_str\}} \\
\texttt{**Error message**: \{error\}} \\
\\
\texttt{\{signature\_info\}} \\
\\
\texttt{**Current retry count**: \{retry\_count\}/\{max\_retries\}} \\
\\
\texttt{Please use the correct parameter names and types, then try again.} \\
\midrule

\textbf{Signature Info Example:} \\
\texttt{**Correct function signature**:} \\
\hspace*{1em}\texttt{add(a, b)} \\
\hspace*{1em}\texttt{Parameters:} \\
\hspace*{2em}\texttt{- a: First number to add} \\
\hspace*{2em}\texttt{- b: Second number to add} \\
\hspace*{1em}\texttt{Description: Add two numbers together} \\
\bottomrule[1.5pt]
\end{tabular}%
\caption{Function call error recovery prompt. Enables LLM self-correction by providing detailed error context and the correct function signature for retry attempts.}
\label{tab:error_recovery_prompt}
\end{table*}


\begin{table*}[t]
\centering
\scriptsize
\renewcommand\arraystretch{1.0}

\begin{tabular}{p{0.95\linewidth}}
\toprule[1.5pt]
\textbf{System Prompt: JSON Format Recovery} \\
\midrule[1pt]

\textbf{Purpose:} When the LLM's response cannot be parsed as valid JSON (e.g., contains extra text, markdown formatting, or syntax errors), this prompt requests a corrected response. The system retries up to 5 times before marking the step as failed. \\
\midrule

\textbf{Prompt Content:} \\
\texttt{Your response could not be parsed as valid JSON. Please respond with ONLY a valid JSON object} \\
\texttt{in this exact format:} \\
\\
\texttt{\{"reasoning": "your thought process", "action": "function\_name", "action\_input": \{"param": value\}\}} \\
\\
\texttt{Do not include any text before or after the JSON.} \\
\bottomrule[1.5pt]
\end{tabular}%
\caption{JSON format recovery prompt. When the LLM outputs malformed JSON, this prompt requests a corrected response in the exact required format, enabling automatic format recovery.}
\label{tab:json_format_recovery}
\end{table*}


\begin{table*}[t]
\centering
\small
\renewcommand\arraystretch{1.2}

\begin{tabular}{p{0.22\linewidth}|p{0.22\linewidth}|p{0.50\linewidth}}
\toprule[1.5pt]
\textbf{Strategy} & \textbf{Phase / Mode} & \textbf{Description} \\
\midrule[1pt]

ReAct & System Prompt & Interleaved reasoning-action loop; outputs JSON with \texttt{reasoning}, \texttt{action}, \texttt{action\_input} \\
\midrule

\multirow{2}{*}{Plan-and-Execute} & Planning & Generate step-by-step plan with \texttt{analysis} and \texttt{steps} \\
\cmidrule{2-3}
 & Execution & Execute steps sequentially; plan injected as context \\
\midrule

\multirow{6}{*}{Execute-Summarize} & Overview & Two-stage: Execute $\rightarrow$ Summarize (Graph Building) \\
\cmidrule{2-3}
 & \multicolumn{2}{l}{\textit{Summarize Stage: Graph Building}} \\
\cmidrule{2-3}
 & \hspace*{1em}ReAct Mode & Iterative graph construction via tool calling \\
\cmidrule{2-3}
 & \hspace*{1em}Plan-and-Build: Planning & Plan graph structure without tool execution \\
\cmidrule{2-3}
 & \hspace*{1em}Plan-and-Build: Execution & Execute plan; plan + trace injected as context \\
\cmidrule{2-3}
 & \hspace*{1em}Input \& Observation & Trace input template; graph state feedback \\
\midrule

\multirow{2}{*}{Error Recovery} & Function Error & Self-correction with correct signature hints \\
\cmidrule{2-3}
 & JSON Format & Format recovery for malformed responses \\

\bottomrule[1.5pt]
\end{tabular}
\caption{Summary of system prompts. ReAct and Plan-and-Execute are base strategies usable independently or within Execute-Summarize. Execute stage: ReAct $|$ Plan-and-Execute. Summarize stage: ReAct $|$ Plan-and-Build.}
\label{tab:prompts_summary}
\end{table*}

\end{document}